\documentclass[11pt,a4paper]{article}
\usepackage{times,latexsym}
\usepackage{url}
\usepackage[T1]{fontenc}

\usepackage[acceptedWithA]{tacl2021v1}

\usepackage{xspace,mfirstuc,tabulary}

\usepackage[utf8]{inputenc}

\usepackage{microtype}

\usepackage{inconsolata}
\usepackage{amsfonts}
\usepackage{graphicx}
\usepackage{amsmath}
\usepackage{array}

\usepackage{inconsolata}

\usepackage{graphicx}
\usepackage{marvosym}  
\usepackage{graphicx}
\usepackage{caption}
\usepackage{stfloats} 
\usepackage{cuted}
\usepackage{multirow}
\usepackage{tabularx}

\usepackage{soul}
\usepackage{tocloft}
\usepackage{etoc}

\usepackage{lipsum}
\usepackage{titletoc}
\usepackage{tcolorbox}
\usepackage{comment}
\usepackage{mathtools}
\usepackage{subcaption}
\usepackage{float}
\usepackage{adjustbox}
\usepackage{enumitem}

\usepackage{booktabs}
\usepackage{dashrule} 
\usepackage{arydshln} 

\newcommand{\blue}[1]{\textcolor{blue}{#1}}

\newcommand{\darkgreen}[1]{\textcolor{green!50!black}{#1}}
\newcommand{\darkred}[1]{\textcolor{red!75!black}{#1}}

\newcommand{\hh}[1]{\textcolor{orange!90!black}{#1}}
\newcommand{\yw}[1]{\textcolor{green!50!black}{#1}}
\newif\iftaclinstructions
\taclinstructionsfalse 
\iftaclinstructions
\renewcommand{\confidential}{}
\renewcommand{\anonsubtext}{(No author info supplied here, for consistency with
TACL-submission anonymization requirements)}
\newcommand{\instr}
\fi

\iftaclpubformat 

\else

\fi

\title{\textit{Beyond One-Size-Fits-All}: Inversion Learning for Highly Effective \\ NLG Evaluation Prompts}

 \author{
    Hanhua Hong$^{1}$\Thanks{Equal contribution}~~,
   Chenghao Xiao$^{2*}$,
   Yang Wang$^1$,
   Yiqi Liu$^1$\\
   \textbf{
   Wenge Rong$^3$,
   Chenghua Lin$^{1}$\Thanks{Corresponding author}~~
   }
   \ \\
   $^1$The University of Manchester~
   $^2$Durham University~
   $^3$Beihang University
   \\
   \texttt{\{hanhua.hong, yang.wang-27\}@postgrad.manchester.ac.uk}
   \\
   \texttt{chenghao.xiao@durham.ac.uk, w.rong@buaa.edu.cn}
   \\
   \texttt{\{yiqi.liu, chenghua.lin\}@manchester.ac.uk}
   \\
 }

\date{}

\begin{document}
\maketitle
\begin{abstract}
Evaluating natural language generation systems is challenging due to the diversity of valid outputs. While human evaluation is the gold standard, it suffers from inconsistencies, lack of standardisation, and demographic biases, limiting reproducibility. 
LLM-based evaluators offer a scalable alternative but are highly sensitive to prompt design, where small variations can lead to significant discrepancies.
In this work, we propose an inversion learning method that learns effective reverse mappings from model outputs back to their input instructions, enabling the automatic generation of highly effective, model-specific evaluation prompts.
Our method requires only a single evaluation sample and eliminates the need for time-consuming manual prompt engineering, thereby improving both efficiency and robustness. Our work contributes toward a new direction for more robust and efficient LLM-based evaluation.

\end{abstract}

\section{Introduction} 
Evaluating natural language generation (NLG) systems is notoriously difficult due to the diversity of valid outputs for a single input~\cite{zhao-etal-2023-evaluating,zhao-etal-2024-slide}~\cite{zhao-etal-2024-slide}. As a result, human assessment remains the most trusted evaluation method. However, despite its importance, the quality of human evaluation is often questioned due to the lack of standardisation, inconsistencies in evaluation executions, and evaluator demographic biases \cite{howcroft-etal-2020-twenty, belz2024inlg, elangovan-etal-2024-considers}. 
\citet{howcroft-etal-2020-twenty} highlight that even after two decades of research, the field still lacks clear definitions and guidelines for key evaluation criteria, making comparisons across studies difficult.

The advent of large language models (LLMs) has introduced a paradigm shift in evaluation, positioning them as surrogate human evaluators. For instance, LLM-based evaluators can process structured prompts to assess multiple aspects of text quality based on explicit criteria (e.g., G-Eval~\cite{liu-etal-2023-g}) or perform comparative judgments between multiple outputs without predefined rubrics (e.g., LLM-as-a-Judge~\cite{zheng2023judging}). 
Their scalability, ability to follow explicit evaluation criteria, and capacity to provide delicate human-like judgments across diverse tasks (e.g., text generation, reasoning, etc.) make them a compelling alternative to both human evaluation and existing automatic metrics such as BERTScore \cite{Zhang2020BERTScore} and BARTScore \cite{NEURIPS2021_e4d2b6e6}, which rely on deterministic similarity measures or generation likelihood estimates \cite{li2024llmsasjudgescomprehensivesurveyllmbased}.

However, LLM-based evaluation also presents inherent challenges, most notably \textit{high sensitivity to prompts}, which, in current practice, are predominantly hand-crafted. 
Extensive literature highlights how prompt design and variations can significantly impact output quality -- even small changes in wording can lead to substantial differences in evaluation results \cite{aher2023using, huijzer2023llm, errica2024did, cao2024worstpromptperformancelarge, sclar2024quantifyinglanguagemodelssensitivity}. 
For instance, subtle variations in few-shot prompt templates have caused performance discrepancies of up to 76 accuracy points on tasks from the Super-Natural Instruction dataset \cite{Polo2024EfficientME}. 
To mitigate this issue, \citet{Polo2024EfficientME} propose estimating the performance distribution across multiple prompt variants, rather than relying on a single prompt for evaluation. \citet{qian-etal-2024-large} benchmark prompts with different components on Machine Translation tasks to figure out which components are crucial for prompt templates.
More recently, \citet{wen-etal-2025-hpss} propose a heuristic search algorithm (HPSS) to navigate a vast combinatorial space of prompt factors. However, such search-based methods are inherently data-hungry and computationally intensive, requiring a large validation dataset to guide prompt optimisation.

To the best of our knowledge, our work presents the first attempt to study the problem of learning high-quality, model-specific evaluation prompts. 
It has been observed that the effectiveness of evaluation guidelines varies among human evaluators \cite{loakman-etal-2023-iron}. In a similar vein, we argue that LLMs from different families (e.g., Qwen \cite{qwen2.5}, LLaMA \cite{grattafiori2024llama3herdmodels}) possess unique characteristics due to their distinct training and alignment techniques \cite{Mu_oz_Ortiz_2024, sun2025idiosyncrasies, lee-etal-2025-llms}. It is therefore reasonable to assume they exhibit different \textit{interpretive biases}, meaning a prompt that is effective for one model may be suboptimal for another.

We tackle this challenge through \textit{inversion learning}, where the core idea is to learn the inverse mapping from evaluation outcomes and inputs back to effective prompts for a given LLM. Our work is fundamentally different from prior studies in two key ways: 
(i) it is the first to explore the automatic generation of evaluation prompts via inversion learning;  
(ii) it requires only a single evaluation sample to generate a highly effective, model-specific prompt.
Specifically, when an LLM acts as an evaluator, we assume that there exists a mapping $f_p(\cdot)$ that maps $X$ (texts to be evaluated) to evaluation outcome $S$, where $S$ approximates the human evaluation distribution $G$. 
We train an inverse model \( \tilde{f} \) to effectively learn the inverse of \( f \), enabling it to generate a model-specific prompt \( p \) given the content to be evaluated \( X \) and a target evaluation outcome \( g \in G \) (e.g., a human-annotated score).

We conducted comprehensive experiments to evaluate the effectiveness of our inverse prompt approach across three key generation tasks -- summarisation, machine translation, and conversational response generation -- using four public datasets and two model families, Qwen~\cite{qwen2.5} and LLaMA~\cite{grattafiori2024llama3herdmodels}, ranging from 3B to 14B parameters. Comparisons against popular human-crafted prompts for these tasks, prompts generated by the original instruction-tuned LLMs, and prompts refined by prompt optimisation methods, demonstrate the superior performance of the evaluation prompts produced by our inverse model. 

To summarise, the contribution of our work is three-fold:
\vspace{5mm}
\setlist{nolistsep}
\begin{itemize}
    \setlength\itemsep{0.6em}
    \item We introduce a one-shot generative paradigm for prompt engineering, making a fundamental shift from the dominant manual or data-intensive, search-based approaches.
    \item We propose a novel inversion learning framework capable of generating model-specific evaluation prompts from a single evaluation sample, eliminating the need for manual or iterative prompt engineering.
    \item We conducted comprehensive experiments across three NLG evaluation tasks, demonstrating that inversely generated prompts are highly effective, consistently outperforming both human-crafted prompts and strong baselines.
\end{itemize}
\vspace{1mm}

Our findings highlight the limitations of relying on manually crafted prompts and underscore the potential of model inversion as a more efficient and scalable approach for high-quality evaluation prompt generation, paving the way for more effective and systematic LLM-based evaluation.

\section{Related Work}

Our work is positioned at the intersection of automatic prompt engineering for LLM-based evaluation and the field of language model inversion. We will discuss each in turn, clarifying how our approach presents a novel contribution.


\begin{figure*}[h!]
    \centering
    \includegraphics[width=\textwidth]{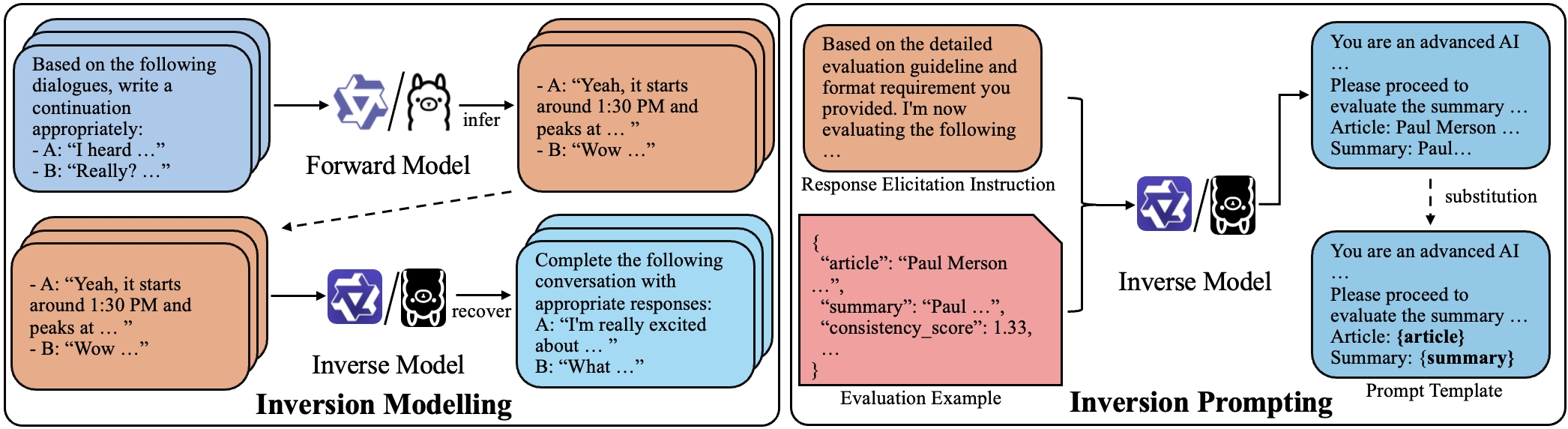}
    \caption{Illustration of the inverse prompt generation process. The bold text in "Prompt Template" indicates substituting the specific example with a generic placeholder.}
    \label{fig:structure_figure}
\end{figure*}

\subsection{Automatic Prompt Engineering for LLM-based Evaluation}

The use of LLMs as evaluators, exemplified by frameworks like G-Eval \cite{liu-etal-2023-g}, has become a scalable alternative to traditional metrics. However, the field has widely acknowledged that the reliability of these evaluators is critically undermined by their sensitivity to the prompts used \cite{sclar2024quantifyinglanguagemodelssensitivity, aher2023using} and by self-preference that can affect evaluation results \cite{liu-etal-2024-llms-narcissistic}. This challenge has catalysed the field of Automatic Prompt Optimisation (APO), which treats prompt design as a search or optimisation problem \cite{ramnath-etal-2025-apo-survey}. 
Gradient-based optimisation methods treat prompts as continuous vectors and use gradient descent to optimise them, but these typically require impractical white-box access to the model's internals \cite{shin-etal-2020-autoprompt, pryzant-etal-2023-automatic}. 
The dominant paradigm in black-box settings is iterative search and inverse-prompting, where an LLM itself acts as an optimiser to iteratively refine candidate prompts based on performance on a validation set \cite{yang-etal-2023-opro, sun2023autohint}. When applied specifically to NLG evaluation, HPSS \cite{wen-etal-2025-hpss} represents the state-of-the-art in this category, employing a heuristic search to find the optimal configuration of prompt components by testing them against a large validation set.

Our work is fundamentally distinct from the existing paradigm for prompt optimisation, where most of the methods are search-based. They start with initial prompts and iteratively refine or select them based on performance on a validation set. Our approach is \textit{generative}. We do not search over the prompt space; rather, we train an inverse model that directly generates a high-quality prompt from a single evaluation sample. The iterative nature of search-based methods makes them inherently data-hungry. HPSS, for instance, requires a large validation set (e.g., using up to 50\% of test data) to guide its search. Our inversion learning approach is exceptionally efficient in its application phase, requiring only a single annotated sample to generate the prompt, thereby eliminating the need for a large validation set for prompt tuning.

\subsection{Language Model Inversion}

Language model inversion reconstructs inputs or instructions from a model’s outputs or internal representations. Early work revealed unintentional memorisation of training data \cite{song2020informationleakageembeddingmodels, Nicholas2021Extracting}, enabling auditing of sensitive information. Subsequent studies showed inversion via next-token distributions \cite{morris2023languagemodelinversion} or black-box output reconstruction \cite{zhang2024extractingpromptsinvertingllm}, indicating that LLMs inherently encode retrievable input traces.

Existing methods typically fall into two categories: output-based inversion infers prior context from next-token probabilities \cite{morris2023languagemodelinversion} or reconstructs prompts from responses \cite{zhang2024extractingpromptsinvertingllm}, but often relies on deterministic decoding and struggles with stochastic sampling strategies such as temperature or nucleus sampling \cite{holtzman2020curiouscaseneuraltext}. 
Embedding-based inversion recovers text from vector embeddings via encoder conditioning \cite{morris2023text} or exploits self-attention gradient structure (DAGER) to reconstruct whole batches exactly \cite{petrov2024dagerexactgradientinversion}, yet typically requires access to model internals. 
To address these gaps, we propose an inversion learning approach for the automatic generation of effective evaluation prompts using a single evaluation sample, offering a more robust, efficient, and adaptable framework for diverse evaluation tasks.

\begin{table*}[!h]
    \centering \small
    \begin{adjustbox}{max width=0.95\textwidth}
    \begin{tabular}{cp{0.65\textwidth}}
        \toprule
        \textbf{Notation} & \textbf{Description} \\
        \midrule
        $\mathcal{D}$ & Instruction tuning dataset used for supervised fine-tuning (SFT) \\
        $x$ & Input prompt in the instruction-tuning dataset \\
        $y$ & Response output paired with instruction prompt $x$ \\
        $\mathcal{M}$ & The original pre-trained model without instruction fine-tuning \\
        $\mathcal{M}_{\mathrm{Instruct}}$ & The instruction-tuned model (aka. the \textit{forward model}) obtained by fine-tuning the base model $\mathcal{M}$ \\
        $\mathcal{M}_{\mathrm{Inverse}}$ & Inverse model obtained by fine-tuning the base model $\mathcal{M}$ on the inverse training dataset $\mathcal{D}_{\mathrm{Inv}}$ \\
        $p(\cdot)$ & Prompt template with a placeholder \\
        $p(c)$ & Fully instantiated evaluation prompt by substituting the placeholder in the prompt template $p(\cdot)$ with content $c$ \\
        \bottomrule
    \end{tabular}
    \end{adjustbox}
    \caption{Summary of notation and terminology used in the methodology section.} 
    \label{tab:term}
\end{table*}


\section{Methodology}
We propose an inversion learning method that learns an effective reverse mapping from model outputs back to their input instructions, enabling the automatic generation of highly effective, model-specific evaluation prompts.
Our approach does not require any  task-specific evaluation data for training and is highly efficient, as it generates an evaluation prompt from a \textit{single data sample} comprising the evaluation content and its corresponding human evaluation result. 
The overall framework includes two stages: inversion modelling and inversion prompting, as shown in Figure~\ref{fig:structure_figure}. 
For completeness, Table~\ref{tab:term} summarises the notation and terminology used throughout this section.




\subsection{Inversion Modelling}
\label{sec:inversion_modelling}

There are two primary settings for training an inverse model: \textit{Black-Box} and \textit{White-Box} Inversion. The black-box setting refers to the scenarios where we do not have access to a model's supervised fine-tuning (SFT) or instruction-tuning data and training process, which is typically the case for most of the existing LLMs. Therefore, in this setting, we approximate the inverse behaviour of publicly available instruction-tuned models without access to their original SFT data. In contrast, the white-box setting assumes full access to both the SFT dataset and the model training pipeline, allowing the inverse model to be trained from scratch using a base pre-trained LLM and a known, controllable data source.   

We primarily focus on black-box inversion, as this setting better reflects realistic deployment scenarios in which models are accessible but not fully transparent (e.g., models with released weights but without access to their training data or full training details). Moreover, off-the-shelf instruction-tuned LLMs typically undergo extensive fine-tuning using carefully curated SFT datasets and reinforcement learning, making them more likely to exhibit strong baseline evaluation capabilities~\cite{zhao2025surveylargelanguagemodels}. 
Nevertheless, we also conduct extensive experiments in the white-box setting to enable controlled comparisons and systematically examine the characteristics of both inversion approaches. 
 
\subsubsection{Black-Box Setting}
\label{sec:black-box}

When training an inverse model in the black-box setting, it is undesirable to simply repurpose an existing instruction-tuning dataset by swapping input and output pairs. This is because the original responses \( y \) in a predefined SFT dataset may not reflect the output characteristics or distribution of the target instruction-tuned model \( \mathcal{M}_{\mathrm{Instruct}} \). Instead, we argue that it is essential to use outputs generated directly by \( \mathcal{M}_{\mathrm{Instruct}} \), as this ensures the inverse model is trained on data that more accurately captures the behavioural patterns of the target model \( \mathcal{M}_{\mathrm{Instruct}} \). Such alignment is crucial for learning effective, model-specific evaluation prompts. To this end, we first perform \textit{inversion dataset distillation}, where model-specific responses \( \tilde{y} \) are generated by performing inference with \( \mathcal{M}_{\mathrm{Instruct}} \) on prompts \( x \) from the SFT dataset.

\noindent\textbf{Inversion Dataset Distillation.}~~
Given a SFT dataset \( \mathcal{D}_{\mathrm{SFT}} = \{(x, y)\} \), where \( x \) represents the input prompts and \( y \) the original target responses, we perform inference using an off-the-shelf instruction-tuned model \( \mathcal{M}_{\mathrm{Instruct}} \) as follows:
\begin{equation}
    \label{inference}
    \tilde{y} = \mathcal{M}_{\mathrm{Instruct}}(x)
\end{equation}

Here \( \tilde{y} \) denotes the model-specific output generated in response to prompt \( x \). We then construct the inversion training dataset \( \mathcal{D}_{\mathrm{Inv}} = \{(\tilde{y}, x)\} \). This inverse dataset serves as the foundation for training the inverse model, which is designed to learn the reverse mapping from model specific outputs back to their corresponding input prompts. 

\noindent\textbf{Inversion-based Fine-tuning.}~~Based on the inverse dataset \( \mathcal{D}_{\mathrm{Inv}} \), we inverse fine-tune a base pre-trained language model \( \mathcal{M} \) (e.g., \texttt{Qwen-2.5}), which shares the same architecture as the instruction-tuned model \( \mathcal{M}_{\mathrm{Instruct}} \) (e.g., \texttt{Qwen-2.5-Instruct}) but has not undergone instruction tuning.
Specifically, we treat the model-generated response \( \tilde{y} \) as the \textit{input} and the original prompt \( x \) as the target \textit{output}, and fine-tune the base model \( \mathcal{M} \) using a standard supervised fine-tuning procedure. 
\begin{equation}
\tilde{\theta} = \underset{\theta}{\arg\min}\, \mathbb{E}_{(\tilde{y}, x) \sim \mathcal{D}_{\mathrm{Inv}}} \Big[ \mathcal{L}\big( \mathcal{M}(\tilde{y};\theta), x \big) \Big]
\end{equation}

This inversion-based fine-tuning process aims to effectively learn to reconstruct the original instruction \( x \) from the corresponding model-generated output \( \tilde{y} \). By capturing the latent correspondence between outputs and their originating instructions, the inverse model internalises the implicit structure of task-specific instructions, thereby enabling the generation of prompts that are more precisely aligned with the behavioural characteristics of the target LLM.

\subsubsection{White-Box Setting}

In contrast to the black-box setting, the white-box setting assumes full control over both the forward and inverse fine-tuning processes. This allows us to fine-tune not only the forward instruction-tuned model but also the inverse model based on the \textit{same} SFT dataset. 

Formally, we begin by training the standard \textit{forward} instruction-tuned model via supervised fine-tuning of a base pre-trained LLM  on a dataset \( \mathcal{D}_{\mathrm{SFT}} = \{(x,y)\} \):
\begin{equation}
\theta_{\text{Instruct}} = \underset{\theta}{\arg\min}\, \mathbb{E}_{(x, y) \sim \mathcal{D}_{\text{SFT}}} \Big[ \mathcal{L}\big(\mathcal{M}(x; \theta), y\big) \Big] 
\end{equation}
where \( \theta_{\text{Instruct}} \) represents the model parameters after instruction tuning. The resulting instruction-tuned model is given as: 
\begin{equation}
\mathcal{M}_{\mathrm{Instruct}}(\cdot) = \mathcal{M}(\cdot;\theta_{\mathrm{Instruct}})
\end{equation}

To train the inverse model, we construct the inversion dataset \( \mathcal{D}_{\mathrm{Inv}} = \{(y,x)\} \) by simply swapping the input–output pairs in \( \mathcal{D}_{\mathrm{SFT}} \), such that the original outputs become inputs and vice versa. The inverse model is subsequently trained using the same SFT procedure as in the Black-Box setting:
\begin{equation}
\tilde{\theta} = \underset{\theta}{\arg\min}\, \mathbb{E}_{(y, x) \sim \mathcal{D}_{\mathrm{Inv}}} \Big[ \mathcal{L}\big(\mathcal{M}(y; \theta), x\big) \Big]
\end{equation}

Finally, the inverse model can be derived as:
\begin{equation}
\mathcal{M}_{\mathrm{Inverse}}(\cdot) = \mathcal{M}(\cdot;\tilde{\theta})
\end{equation}

\subsection{Inversion Prompting}
\label{sec:inversion_prompting}

\begin{figure}[t]
\centering

\tcbset{
    colback=gray!5!white,
    fontupper=\scriptsize,
    width=0.95\columnwidth,
    left=1pt,
    right=1pt
}
\begin{tcolorbox}[]
\textcolor{blue}{Based on the detailed evaluation guideline and format requirement you provided, I'm now evaluating consistency of the following summary to the article with a score between 0 and 1:\\}
```json\\
\{\\
\textit{"article":"A woman \dots}",\\
\textit{"summary":"The mother \dots}",\\
\textbf{"consistency\_score": 0.66666}\\
\}
'''
\end{tcolorbox}
\caption{Example of a meta-prompt for the inverse model.}
\label{fig:template_to_inversion_model}
\end{figure}

Upon training, the inverse model is expected to have learned an effective reverse mapping from model-specific outputs back to their corresponding input instructions, ultimately enabling the generation of effective evaluation prompts tailored to the target instruction-tuned LLM. 

To generate evaluation prompts, we feed the inverse model a meta-prompt input \( I_t = (E_{\mathcal{T}}, c_t, r_t) \), which consists of three components: the evaluation content \( c_t \), the corresponding human evaluation result \( r_t \), and a response elicitation instruction \( E_{\mathcal{T}} \). We adopt a \textit{one-shot strategy}, where a single data pair \( (c_t, r_t) \) is randomly sampled from a downstream evaluation task \( \mathcal{T} = \{(c, r)\} \), with \( c \) denoting the content to be evaluated (e.g., a translation and its source sentence) and \( r \) the associated human judgment. 
The response elicitation instruction is designed to guide the model to generate prompts that are both well-structured and aligned with the evaluation aspects reflected in the result (e.g., consistency). It also encourages the model to output structured evaluative instructions rather than free-form explanations or responses. 
Figure~\ref{fig:template_to_inversion_model} illustrates such an input example, where the evaluation content is shown in \textit{italics}, the result in \textbf{bold}, and the response elicitation instruction \( E_{\mathcal{T}} \) in \textcolor{blue}{blue}.

The inverse prompt generation process is formally defined as:
\begin{equation}
    \label{inverse_prompt}
    p(c_t) = \mathcal{M}_{\mathrm{Inverse}}\big(I_t\big)
\end{equation}
The objective is to produce an evaluation prompt \( p(c_t) \) such that, when used by the evaluator \( \mathcal{M}_{\mathrm{Instruct}} \), it yields an evaluation outcome that closely approximates the human-provided evaluation result \( r_t \).

The inversion-generated prompt \( p(c_t) \) typically includes \( c_t \), the original content to be evaluated, as illustrated in Figure~\ref{fig:structure_figure}. This is expected, as the generated evaluation prompt is designed to assess a specific input text. To construct a generalisable evaluation prompt template \( p(\cdot) \), we automatically replace the content in \( p(c_t) \) that is specific to the one-shot example with format placeholders. Once the general evaluation prompt template is obtained, it can be used to evaluate any input from the same downstream task by infilling the template with the target evaluation content and passing it to the corresponding forward instruction-tuned model. 
Note that it is essential to use the corresponding forward model as the evaluator, rather than the inverse model, since the inversion training process optimises the model for generating evaluation prompts rather than for performing the actual evaluation.
Given template \( p(\cdot) \), the predicted evaluation outcome \( \hat{r}_i \) for any \( c_i \) are computed as:
\begin{equation}
    \label{evaluation}
    \hat{r}_i = \mathcal{M}_{\text{Instruct}} \big( p(c_i) \big)
\end{equation}

The transformation from an instance-specific prompt to general template is visually highlighted in bold in Figure~\ref{fig:structure_figure}.
Examples of the one-shot input format and the corresponding inverse model outputs for various datasets are provided in Appendix~\ref{app_output}.

\section{Experimental Setup}

\begin{table*}[h!]
    \centering
    \begin{adjustbox}{max width=0.95\textwidth}
    \begin{tabular}{lcccccccccccc}
        \toprule
        \textbf{\texttt{Evaluator}} & \multicolumn{2}{c}{\textbf{SummEval}} & \multicolumn{2}{c}{\textbf{QAGS-C}} & \multicolumn{2}{c}{\textbf{QAGS-X}} & \multicolumn{2}{c}{\textbf{Topical-Chat}} & \multicolumn{2}{c}{\textbf{WMT-22}} & \multicolumn{2}{c}{\textbf{Average}} \\
        \cmidrule(lr){2-3}\cmidrule(lr){4-5}\cmidrule(lr){6-7}\cmidrule(lr){8-9}\cmidrule(lr){10-11}\cmidrule(lr){12-13}
        \textit{~Evaluation Prompt}  & $\rho$ & $r$ & $\rho$ & $r$ & $\rho$ & $r$ & $\rho$ & $r$ & $\rho$ & $r$ & $\rho$ & $r$\\
        \midrule
        \texttt{BERTScore} & 0.290 & 0.317 & 0.505 & 0.576 & 0.008 & 0.024 & 0.273 & 0.262 & 0.277 & 0.328 & 0.271 & 0.301\\
        \texttt{BARTScore} & 0.385 & 0.414 & 0.680 & 0.735 & 0.159 & 0.184 & 0.119 & 0.138 & 0.202 & 0.128 & 0.309 & 0.320 \\
        \texttt{HPSS(Qwen-2.5-7B-Instruct) } \\
        \textit{~\textbf{10}-shot} & 0.373 & 0.402 & 0.514 & 0.496 & 0.164 & 0.048 & 0.465   & 0.232 & 0.161 & 0.040 & 0.335 & 0.296\\
        \textit{~\textbf{30}-shot} & 0.403 & 0.423 & 0.530 & 0.522 & 0.354 & 0.382 & 0.478 & 0.420 & 0.198 & 0.242 & 0.392 & 0.398\\
        \textit{~\textbf{50}-shot} & 0.406 & 0.444 & 0.563 & 0.608 & 0.360 & 0.332 & 0.483 & 0.474 & 0.252 & 0.218 & 0.413 & 0.415\\
        \textit{~\textbf{100}-shot} & 0.429 & 0.450 & 0.582 & 0.577 & 0.461 & 0.442 & 0.503 & 0.500 & 0.240 & 0.259 & 0.443 & 0.445\\
        
        \midrule
        \multicolumn{3}{l}{\texttt{LLaMA-3.1-8B-Instruct}} & \multicolumn{8}{c}{\textbf{BlackBox Setting}} \\
        \textit{~Human-Crafted Prompt} & \underline{0.375} & \underline{0.433} & \underline{0.558} & \underline{0.590} & \underline{0.376} & \underline{0.350} & 0.385 & 0.372 & \underline{0.259} & \textbf{0.292} & \underline{0.391} & \underline{0.407}\\
        \textit{~Forward Prompt} & 0.268 & 0.286 & 0.531 & 0.569 & 0.137 & 0.126 & \underline{0.419} & \underline{0.407} & 0.233 & 0.248 & 0.318 & 0.327\\
        \textit{~Inverse Prompt (\textbf{Ours})} & \textbf{0.400} & \textbf{0.466} & \textbf{0.598} & \textbf{0.620} & \textbf{0.405} & \textbf{0.401} & \textbf{0.437} & \textbf{0.423} & \textbf{0.277} & \underline{0.256} & \textbf{0.423} & \textbf{0.433}\\
        Relative Gain & \darkgreen{$\uparrow $49\%} & \darkgreen{$\uparrow $63\%} & \darkgreen{$\uparrow $13\%} & \darkgreen{$\uparrow $9\%} & \darkgreen{$\uparrow $196\%} & \darkgreen{$\uparrow $218\%} & \darkgreen{$\uparrow $4\%} & \darkgreen{$\uparrow $4\%} & \darkgreen{$\uparrow $19\%} & \darkgreen{$\uparrow $3\%} & \darkgreen{$\uparrow $33\%} & \darkgreen{$\uparrow $32\%}\\
        \noalign{\vspace{0.4ex}}         \hdashline         \noalign{\vspace{0.65ex}} 
        \texttt{Qwen-2.5-7B-Instruct} &&&&&&&&&&&& \\
        \textit{~Human-Crafted Prompt} & \underline{0.374} & \underline{0.430} & \underline{0.654} & \underline{0.668} & \underline{0.483} & \underline{0.464} & 0.398 & 0.393 & 0.271 & 0.202 & \underline{0.436} & \underline{0.431}\\
        \textit{~Forward Prompt} & 0.315 & 0.339 & 0.529 & 0.603 & 0.198 & 0.207 & \underline{0.436} & \underline{0.439} & \underline{0.274} & \underline{0.284} & 0.350 & 0.374\\
        \textit{~Inverse Prompt (\textbf{Ours})} & \textbf{0.418} & \textbf{0.457} & \textbf{0.661} & \textbf{0.673} & \textbf{0.524} & \textbf{0.530} & \textbf{0.502} & \textbf{0.501} & \textbf{0.313} & \textbf{0.316} & \textbf{0.484} & \textbf{0.495}\\
        Relative Gain & \darkgreen{$\uparrow$ 33\%} & \darkgreen{$\uparrow $35\%} & \darkgreen{$\uparrow $25\%} & \darkgreen{$\uparrow $12\%} & \darkgreen{$\uparrow $164\%} & \darkgreen{$\uparrow $156\%} & \darkgreen{$\uparrow $15\%} & \darkgreen{$\uparrow $14\%} & \darkgreen{$\uparrow $11\%} & \darkgreen{$\uparrow $11\%} & \darkgreen{$\uparrow $38\%} & \darkgreen{$\uparrow $32\%}\\
        \midrule
        \multicolumn{3}{l}{\texttt{LLaMA-3.1-8B-WhiteBox}} & \multicolumn{8}{c}{\textbf{WhiteBox Setting}} \\
        \textit{~Human-Crafted Prompt} & \underline{0.341} & \underline{0.392} & \underline{0.555} & \underline{0.542} & \underline{0.254} & \underline{0.254} & \textbf{0.446} & \textbf{0.436} & \textbf{0.274} & \textbf{0.274} & \underline{0.374} & \underline{0.380}\\
        \textit{~Forward Prompt} & 0.334 & 0.374 & 0.444 & 0.491 & 0.170 & 0.166 & 0.318 & 0.300 & 0.249 & 0.250 & 0.303 & 0.318\\
        \textit{~Inverse Prompt (\textbf{Ours})} & \textbf{0.388} & \textbf{0.440} & \textbf{0.576} & \textbf{0.577} & \textbf{0.356} & \textbf{0.336} & \underline{0.441} & \underline{0.422} & \underline{0.257} & \underline{0.256} & \textbf{0.404} & \textbf{0.406}\\
        Relative Gain & \darkgreen{$\uparrow $16\%} & \darkgreen{$\uparrow $18\%} & \darkgreen{$\uparrow $30\%} & \darkgreen{$\uparrow $18\%} & \darkgreen{$\uparrow $109\%} & \darkgreen{$\uparrow $102\%} & \darkgreen{$\uparrow $39\%} & \darkgreen{$\uparrow $40\%} & \darkgreen{$\uparrow $3\%} & \darkgreen{$\uparrow $2\%} & \darkgreen{$\uparrow $33\%} & \darkgreen{$\uparrow $28\%}\\
        \noalign{\vspace{0.4ex}} 
        \hdashline
        \noalign{\vspace{0.65ex}} 
        \texttt{Qwen-2.5-7B-WhiteBox} &&&&&&&&&&&& \\
        \textit{~Human-Crafted Prompt} & \textbf{0.406} & \textbf{0.467} & \underline{0.557} & \underline{0.604} & \underline{0.416} & \underline{0.410} & \underline{0.427} & \underline{0.420} & \underline{0.292} & \underline{0.259} & \underline{0.420} & \underline{0.432}\\
        \textit{~Forward Prompt} & 0.341 & 0.360 & 0.236 & 0.269 & 0.321 & 0.335 & 0..419 & 0.416 & 0.251 & 0.246 & 0.314 & 0.325\\
        \textit{~Inverse Prompt (\textbf{Ours})} & \underline{0.402} & \underline{0.425} & \textbf{0.661} & \textbf{0.669} & \textbf{0.590} & \textbf{0.602} & \textbf{0.464} & \textbf{0.461} & \textbf{0.301} & \textbf{0.286} & \textbf{0.484} & \textbf{0.489}\\
        Relative Gain & \darkgreen{$\uparrow $20\%} & \darkgreen{$\uparrow $14\%} & \darkgreen{$\uparrow $49\%} & \darkgreen{$\uparrow $36\%} & \darkgreen{$\uparrow $247\%} & \darkgreen{$\uparrow $262\%} & \darkgreen{$\uparrow $46\%} & \darkgreen{$\uparrow $54\%} & \darkgreen{$\uparrow $21\%} & \darkgreen{$\uparrow $14\%} & \darkgreen{$\uparrow $60\%} & \darkgreen{$\uparrow $54\%}\\
        \bottomrule
    \end{tabular}
    \end{adjustbox}
    \caption{Results of average Spearman ($\rho$) and Pearson($r$) correlations on various datasets with different models and settings. The \texttt{LLaMA-3.1-8B-WhiteBox} and \texttt{Qwen-2.5-7B-WhiteBox} are models instruction-tuned by us with the Infinity-Instruct dataset in the white-box setting. \textbf{QAGS-C} denotes QAGS-CNN dataset, and \textbf{QAGS-X} denotes QAGS-XSUM dataset. Within the data of the same model, the bold values indicate the best results and the underscored values indicate the second-best ones. \textit{Relative Gain} denotes the increase rate of the performance of Inverse Prompt to that of the corresponding Forward Prompt.}
    \label{tab:main_result}
\end{table*}

\noindent\textbf{Data.}~~In our experiments, we select Infinity-Instruct \cite{li2025infinityinstructscalinginstruction}, a large-scale instruction-following dataset, as the SFT dataset. 
Our considerations for selecting this dataset are two-fold: 
(i) \underline{\textit{Quality}}: when fine-tuning widely adopted base models on Infinity-Instruct, they achieve state-of-the-art results without requiring reinforcement learning from human feedback (RLHF)~\cite{xie2025infir}. This underscores the dataset's exceptional quality compared to alternatives;  
(ii) \underline{\textit{Diversity}}: the dataset spans over 20 diverse domains, enabling the inverse model to learn a generalised inverse mapping and support robust performance even on tasks not explicitly present in the training data.

Due to resource constraints, we select the 0625 subset of Infinity-Instruct, which contains approximately 660k samples. 
In the white-box setting, both the forward instruction-tuned model and the inverse model are trained using this subset. 
In the black-box setting, we first construct the inverse training dataset by distilling from the corresponding instruction-tuned LLM using only the inputs from Infinity-Instruct. 
We then pair the output with the corresponding input to form the inverse training set, where the input-output pairs are reversed (see \S\ref{sec:black-box}). 
This inverse dataset is subsequently used to train the inverse model from the \textit{base version} of the same LLM.

\noindent\textbf{Evaluation Protocol.}~~We conduct experiments on three prominent text generation tasks: summarisation, conversational response generation, and machine translation. 
Following standard practice \cite{zhong2022unifiedmultidimensionalevaluatortext,gao2025analyzingevaluatingcorrelationmeasures}, we assess the performance of LLM-based evaluators by calculating Spearman ($\rho$) and Pearson ($r$) correlations between the evaluator's predicted scores and the ground truth scores annotated by humans. 
Following G-EVAL \cite{liu-etal-2023-g}, we select SummEval \cite{fabbri-etal-2021-summeval}, Question Answering and Generation for Summarization (QAGS, \citealp{wang-etal-2020-asking}), and Topical-Chat \cite{Gopalakrishnan2019TopicalChatTK} as benchmarks for summarisation and response generation. QAGS consists of two subtasks: QAGS-CNN and QAGS-XSUM, with the latter containing more abstract summaries.
For machine translation, we use the English-to-German corpus constructed by \citet{qian-etal-2024-large}, sourced from WMT-22 \cite{freitag-etal-2022-results}.

\noindent\textbf{Baselines.}~~We compare the effectiveness of prompts generated by our inverse models against three baselines: (i) popular human-crafted prompts for each task, (ii) prompts generated directly by the corresponding forward instruction-tuned LLMs, and (iii) prompts produced by prompt optimisation techniques.
For summarisation and chat response generation, we use human-crafted prompts from Five-stars \cite{wang-etal-2023-chatgpt} and G-EVAL \cite{liu-etal-2023-g}. For machine translation, we use GEMBA \cite{kocmi-federmann-2023-large} and an enhanced version of GEMBA incorporating evaluation guidelines (GEMBA+) \cite{qian-etal-2024-large}. Additionally, we use prompts from  Direct Assessment (DA) across all three tasks. 
We report the best result among all human-crafted prompts for each task. For the inverse prompt, it is generated based on a one-shot example randomly sampled from the corresponding training set. Furthermore, we compare our method against a recent strong baseline, HPSS \cite{wen-etal-2025-hpss}, a prompt optimisation approach. In contrast to our one-shot inversion method, HPSS requires a substantial validation set for tuning; we experiment with validation sizes ranging from 10 to 100 samples.

We also attempt to benchmark against a recent work \texttt{output2prompt} \cite{zhang2024extractingpromptsinvertingllm} using our settings. However, \texttt{output2prompt} failed to generate meaningful prompts for evaluation tasks, instead producing nonsensical repetitions. Consequently, we excluded it from our experiments.

\noindent\textbf{Environment.}~~All experiments and training are conducted using four NVIDIA A100-SXM-80GB GPUs with LoRA~\cite{hu2022lora}, based on the LLaMA-Factory framework~\cite{zheng2024llamafactory}. Please refer to Appendix~\ref{app_environment} for more details.

\section{Experimental Results}

\subsection{Overall Results}

Table~\ref{tab:main_result} presents the main experimental results under both the Black-Box and White-Box settings. To evaluate the generalisability of our approach, we conduct experiments using two prominent open-source LLM families: LLaMA and Qwen.

\noindent\textbf{Black-Box Setting.}~~In the black-box setting, our inverse prompts consistently achieve superior alignment with human judgments compared to both the forward and human-crafted prompts across all datasets and models. 
For instance, on \texttt{LLaMA-3.1-8B-Instruct},
the inverse prompt improves over forward prompts with substantial gains of 33\% in average Spearman correlation ($\rho$) and 32\% in Pearson correlation ($r$), suggesting that standard forward instruction-tuned models are suboptimal for generating effective evaluation prompts.
A similar trend is observed with \texttt{Qwen-2.5-7B-Instruct}, where the inverse prompt achieves the highest average correlation scores ($\rho = 0.484$, $r = 0.495$), outperforming the human-crafted and forward prompts over 13\% and 35\% on average, respectively. 

Compared to the prompt optimisation method HPSS, our inverse prompts achieve significantly higher evaluation performance, despite using only a single annotated example.
Using Qwen-2.5-7B-Instruct as the evaluator, the inverse prompt generated by our Qwen-based inverse model outperforms HPSS prompts across all tested tasks.
On average, our inverse prompts outperform HPSS's 10-shot prompts by 56\% in terms of mean Spearman and Pearson correlation.
Even when HPSS is provided with 100-shot validation data, our method still achieves a 10\% performance advantage. This makes our inversion-based prompt generation approach significantly more practical and data-efficient, especially for evaluating new generation tasks where labelled data is scarce.


\noindent\textbf{White-Box Setting.}~~In this setting, we train both the forward and inverse models using the same SFT dataset. Inverse prompts again, yield the best overall performance, followed by human-crafted prompts.
Notably, inversion learning under the black-box setting achieves higher performance than the white-box setting for LLaMA, while showing similar trends for Qwen. We hypothesise that these patterns may arise from differences in the capabilities of the underlying forward models. 
The instruction-tuned LLMs used in the black-box setting (i.e., \texttt{LLaMA-Instruct} and \texttt{Qwen-Instruct}) have undergone extensive fine-tuning on large-scale, high-quality supervised datasets, followed by additional stages such as reinforcement learning and post-training refinements~\cite{rafailov2023direct}. In contrast, the white-box models trained in our experiments were fine-tuned solely through supervised fine-tuning on a relatively small dataset of around 660k samples. Therefore, while inversion learning achieves better performance in the black-box setting, it remains inconclusive which setup is more optimal, as differences in training strategy and data scale confound a strict comparison.

\begin{table*}[h!]
    \centering
    \newcolumntype{g}{>{\columncolor{lightgray}}c}
    \begin{adjustbox}{max width=0.95\textwidth}
    \begin{tabular}{lcccccccccccc}
        \toprule
        \textbf{\texttt{Evaluator}} & \multicolumn{2}{c}{\textbf{SummEval}} & \multicolumn{2}{c}{\textbf{QAGS-C}} & \multicolumn{2}{c}{\textbf{QAGS-X}} & \multicolumn{2}{c}{\textbf{Topical-Chat}} & \multicolumn{2}{c}{\textbf{WMT-22}} & \multicolumn{2}{c}{\textbf{Average}} \\
        \cmidrule(lr){2-3}\cmidrule(lr){4-5}\cmidrule(lr){6-7}\cmidrule(lr){8-9}\cmidrule(lr){10-11}\cmidrule(lr){12-13}
        \textit{~Evaluation Prompt} & $\rho$ & $r$ & $\rho$ & $r$ & $\rho$ & $r$ & $\rho$ & $r$ & $\rho$ & $r$ & $\rho$ & $r$\\
        \midrule
        \texttt{LLaMA-3.1-8B-Instruct} &&&&&&&&&&&& \\
        \textit{~Forward Prompt} & 0.268 & 0.286 & 0.531 & 0.569 & 0.137 & 0.126 & \underline{0.419} & \underline{0.407} & 0.233 & 0.248 & 0.318 & 0.327\\
        \textit{~Inverse Prompt-Qwen} & 0.376 & 0.434 & 0.569 & 0.607 & 0.305 & 0.317 & 0.400 & 0.385 & 
        \underline{0.271} & \underline{0.254} & 0.384 & 0.399\\
        \textit{~Inverse Prompt-WB} & \textbf{0.411} & \underline{0.443} & \textbf{0.626} & \textbf{0.636} & \underline{0.381} & \underline{0.332} & 0.417 & 0.405 & 0.263 & 0.211 & \underline{0.420} & \underline{0.405}\\
        \textit{~Inverse Prompt (\textbf{Ours})} & \underline{0.400} & \textbf{0.466} & \underline{0.598} & \underline{0.620} & \textbf{0.405} & \textbf{0.401} & \textbf{0.437} & \textbf{0.423} & \textbf{0.277} & \textbf{0.256} & \textbf{0.423} & \textbf{0.433}\\
        \noalign{\vspace{0.4ex}} 
        \hdashline
        \noalign{\vspace{0.65ex}} 
        \texttt{Qwen-2.5-7B-Instruct} &&&&&&&&&&&& \\
        \textit{~Forward Prompt} & 0.315 & 0.339 & 0.529 & 0.603 & 0.198 & 0.207 & 0.436 & 0.439 & 0.274 & 0.284 & 0.350 & 0.374\\
        \textit{~Inverse Prompt-LLaMA} & \underline{0.391} & \underline{0.426} & 0.624 & 0.672 & 0.499 & 0.470 & 0.461 & 0.452 & 0.263 & 0.270 & 0.448 & 0.458\\
        \textit{~Inverse Prompt-WB} & 0.360 & 0.403 & \underline{0.631} & \underline{0.673} & \underline{0.507} & \underline{0.493} & \underline{0.479} & \underline{0.459} & \underline{0.304} & \underline{0.285} & \underline{0.456} & \underline{0.463}\\
        \textit{~Inverse Prompt (\textbf{Ours})} & \textbf{0.418} & \textbf{0.457} & \textbf{0.661} & \textbf{0.673} & \textbf{0.524} & \textbf{0.530} & \textbf{0.502} & \textbf{0.501} & \textbf{0.313} & \textbf{0.316} & \textbf{0.484} & \textbf{0.495}\\ 
        \bottomrule
    \end{tabular}
    \end{adjustbox}
    \caption{Results of the prompt swapping experiment in the black-box setting. Inverse Prompt-Qwen and Inverse Prompt-LLaMA denotes swapping prompts with another model. Inverse Prompt-WB denotes using inverse prompts generated by white-box (WB) models.}
    \label{tab:further_analysis}
\end{table*}
\begin{table*}[h!]
    \centering
    \newcolumntype{g}{>{\columncolor{lightgray}}c}
    \begin{adjustbox}{max width=0.95\textwidth}
    \begin{tabular}{lcccccccccccc}
        \toprule
        \textbf{\texttt{Evaluator}} & \multicolumn{2}{c}{\textbf{SummEval}} & \multicolumn{2}{c}{\textbf{QAGS-C}} & \multicolumn{2}{c}{\textbf{QAGS-X}} & \multicolumn{2}{c}{\textbf{Topical-Chat}} & \multicolumn{2}{c}{\textbf{WMT-22}} & \multicolumn{2}{c}{\textbf{Average}} \\
        \cmidrule(lr){2-3}\cmidrule(lr){4-5}\cmidrule(lr){6-7}\cmidrule(lr){8-9}\cmidrule(lr){10-11}\cmidrule(lr){12-13}
        \textit{~Evaluation Prompt} & $\rho$ & $r$ & $\rho$ & $r$ & $\rho$ & $r$ & $\rho$ & $r$ & $\rho$ & $r$ & $\rho$ & $r$\\
        \midrule
        \texttt{GPT-4o-mini} &&&&&&&&&&&& \\
        \textit{~Forward Prompt-BB} & 0.432 & 0.457 & 0.585 & 0.630 & 0.311 & 0.301 & 0.553 & 0.546 & 0.291 & 0.320 & 0.434 & 0.451\\
        \textit{~Inverse Prompt-BB} & 0.476 & 0.517 & 0.738 & 0.766 & 0.568 & 0.600 & 0.578 & 0.572 & 0.307 & 0.323 & 0.533 & 0.556\\
        Relative Gain & \darkgreen{$\uparrow $10\%} & \darkgreen{$\uparrow $13\%} & \darkgreen{$\uparrow $26\%} & \darkgreen{$\uparrow $22\%} & \darkgreen{$\uparrow $83\%} & \darkgreen{$\uparrow $99\%} & \darkgreen{$\uparrow $5\%} & \darkgreen{$\uparrow $5\%} & \darkgreen{$\uparrow $5\%} & \darkgreen{$\uparrow $1\%} & \darkgreen{$\uparrow $23\%} & \darkgreen{$\uparrow $23\%}\\
        \noalign{\vspace{0.4ex}} 
        \hdashline
        \noalign{\vspace{0.65ex}} 
        \textit{~Forward Prompt-WB} & 0.403 &  0.427 & 0.582 & 0.622 & 0.515 & 0.487 & 0.555 & 0.550 & 0.287 & 0.312 & 0.468 & 0.480\\
        \textit{~Inverse Prompt-WB} & 0.439 & 0.479 & 0.694 & 0.672 & 0.561 & 0.571 & 0.548 & 0.540 & 0.305 & 0.350 & 0.509 & 0.522\\
        Relative Gain & \darkgreen{$\uparrow $9\%} & \darkgreen{$\uparrow $12\%} & \darkgreen{$\uparrow $19\%} & \darkgreen{$\uparrow $8\%} & \darkgreen{$\uparrow $9\%} & \darkgreen{$\uparrow $17\%} & \darkred{$\downarrow$1\%} & \darkred{$\downarrow$2\%} & \darkgreen{$\uparrow $6\%} & \darkgreen{$\uparrow $12\%} & \darkgreen{$\uparrow $9\%} & \darkgreen{$\uparrow $9\%}\\
        \bottomrule
    \end{tabular}
    \end{adjustbox}
    \caption{Results of applying inversion and forward prompts generated by \texttt{Qwen-2.5-7B} models to \texttt{GPT-4o-mini}.}
    \label{tab:gpt-4o_results}
\end{table*}

Task-wise, the effectiveness of inverse prompts is especially pronounced on the QAGS-XSUM dataset, where we observe an average gain of over 100\% to 250\% compared to forward prompts, followed by SummEval with an average gain of 31\%.  
In contrast, machine translation tasks exhibit the smallest relative gains, particularly for LLaMA, which aligns with previous observations~\cite{leiter-eger-2024-prexme}. 
A plausible explanation lies in the nature of the tasks: summarisation tasks are inherently more complex and abstract, and typically exhibit greater variability than translation tasks. Consequently, inversion-generated prompts are likely to yield substantially larger performance improvements for summarisation compared to machine translation.

Overall, these findings demonstrate that inversion learning can effectively generate model-specific evaluation prompts that outperform human-crafted prompts, forward prompts, and strong prompt optimisation methods, yielding assessments that align more closely with human judgments across diverse generation tasks.

\subsection{Sensitivity Analysis of Inverse Prompts}

\noindent\textbf{Model Sensitivity.}~~One of our main hypotheses is that to maximise the effectiveness of an evaluation prompt, it needs to be tailored to the specific LLM. 
In other words, a prompt that is highly effective for one model may perform suboptimally on another.
To investigate this, we design a \textit{prompt swapping} experiment to evaluate the performance of inverse prompts generated by one model family when applied to another (e.g., prompts generated by \texttt{inverse-Qwen} and used by \texttt{forward-LLaMA}). Additionally, since we explore two inverse model training strategies, we also test the cross-strategy effectiveness of prompts, i.e., applying prompts generated by a white-box model on a black-box evaluator.

As shown in Table~\ref{tab:further_analysis}, applying inverse prompts generated by a different model family leads to a noticeable drop in evaluation performance. For example, when prompts generated by \texttt{inverse-Qwen} are applied to \texttt{LLaMA-3.1-8B-Instruct} as the evaluator, the average Spearman and Pearson correlations drop from 0.423 to 0.384 and from 0.433 to 0.399, respectively, compared to when the same prompts are applied to \texttt{Qwen-2.5-7B-Instruct}. A similar performance drop is observed when using \texttt{Qwen-2.5-7B-Instruct} as the evaluator with inverse prompts generated by LLaMA. These findings demonstrate that prompts transferred across different evaluator models significantly lose their effectiveness, highlighting the necessity of generating model-specific inverse prompts. 
When examining cross-strategy sensitivity (i.e., using prompts generated by white-box models on black-box evaluators), we also observe a performance degradation, although the impact is generally less severe than that observed in cross-model transfers. 

Since LLaMA and Qwen are open-source models, we further investigate the sensitivity of inverse prompts generated by \texttt{inverse-Qwen} under both black-box and white-box settings by applying them to the proprietary model \texttt{GPT-4o-mini}, as shown in Table~\ref{tab:gpt-4o_results}.
The results show that inverse prompts continue to outperform forward prompts, with a performance difference of 23\% in the black-box setting. This further demonstrates that our inverse model is capable of generating higher-quality and more effective evaluation prompts than those produced by standard forward instruction-tuned models. 
Nevertheless, the performance gains are less significant compared to when the inverse prompts are applied to the specific model family from which they were generated  (cf. Table~\ref{tab:main_result}). 
Additionally, using \texttt{GPT-4o-mini} as the evaluator yields higher overall performance compared to LLaMA-8B and Qwen-7B, which is unsurprising given its stronger underlying capabilities.
However, using a larger Qwen-14B model can actually surpass \texttt{GPT-4o-mini} in performance. See \S\ref{subsec:model_size_scaling} for a detailed discussion.

Overall, the above analysis reinforces our hypothesis that inverse prompts are most effective when tailored to the specific LLM, and that the prevailing community practice of using \textit{one-size-fits-all} evaluation prompts is sub-optimal. We therefore advocate for the use of model-specific prompts for more accurate and reliable prompt-based evaluation with LLMs.

\begin{table*}[h!]
    \centering
    \begin{adjustbox}{max width=0.95\textwidth}
    \begin{tabular}{lcccccccccccc}
        \toprule
        \textbf{\texttt{Evaluator}} & \multicolumn{2}{c}{\textbf{SummEval}} & \multicolumn{2}{c}{\textbf{QAGS-C}} & \multicolumn{2}{c}{\textbf{QAGS-X}} & \multicolumn{2}{c}{\textbf{Topical-Chat}} & \multicolumn{2}{c}{\textbf{WMT-22}} & \multicolumn{2}{c}{\textbf{Average}}\\
        \cmidrule(lr){2-3}\cmidrule(lr){4-5}\cmidrule(lr){6-7}\cmidrule(lr){8-9}\cmidrule(lr){10-11}\cmidrule(lr){12-13}
        \textit{~Evaluation Prompt} & $\rho$ & $r$ & $\rho$ & $r$ & $\rho$ & $r$ &  $\rho$ & $r$ & $\rho$ & $r$ & $\rho$ & $r$ \\
        \midrule
        \texttt{Qwen-2.5-7B-Instruct} \\
        \textit{~Inverse Prompt} & 0.418 & 0.457 & \underline{0.661} & \textbf{0.673} & \textbf{0.524} & \textbf{0.530} & 0.502 & 0.501 & \textbf{0.313} & \textbf{0.316} & \textbf{0.484} & \textbf{0.495}\\
        \textit{~One Decimal Place} & \underline{0.423} & \underline{0.469} & 0.635 & 0.624 & \underline{0.495} & \underline{0.486} & \underline{0.532} & \textbf{0.515} & 0.301 & 0.264 & \underline{0.477} & 0.472\\
        \textit{~w/o Score Range} & \textbf{0.427} & \textbf{0.484} & 0.652 & \underline{0.669} & 0.422 & 0.429 & 0.516 & 0.493 & 0.301 & \underline{0.314} & 0.464 & \underline{0.478}\\
        \textit{~w/o Score} & 0.405 & 0.446 & \textbf{0.662} & 0.658 & 0.437 & 0.384 & \textbf{0.533} & \underline{0.515} & \underline{0.306} & 0.304 & 0.469 & 0.461 \\
        \bottomrule
    \end{tabular}
    \end{adjustbox}
    \caption{Model sensitivity ablation study.}
    \label{tab:ablation_study}
\end{table*}

\begin{figure*}[!h]
\centering
\tcbset{
    colback=gray!5!white,
    fontupper=\scriptsize,
    left=1pt,
    right=1pt,
    valign=center,
    height=.21\textheight
}
\begin{minipage}[c]{.24\textwidth}
\centering
\begin{tcolorbox}[]
Based on the detailed evaluation guideline and format requirement you provided, I'm now evaluating consistency of the following summary to the article with a score \blue{between 0 and 1}:

```json\\
\{\\
  "article":"A woman \dots",\\
  "summary":"The mother \dots",\\
  "consistency\_score": \blue{0.66666}\\
\}\\
```~~~~~~~~~~~~~~~~~~~(a)
\end{tcolorbox}
\end{minipage}
\begin{minipage}[c]{.24\textwidth}
\centering
\begin{tcolorbox}[]
Based on the detailed evaluation guideline and format requirement you provided, I'm now evaluating consistency of the following summary to the article with a score \blue{between 0 and 1}:

```json\\
\{\\
  "article":"A woman \dots",\\
  "summary":"The mother \dots",\\
  "consistency\_score": \blue{0.7}\\
\}\\
```~~~~~~~~~~~~~~~~~~~(b)
\end{tcolorbox}
\end{minipage}
\begin{minipage}[c]{.24\textwidth}
\centering
\begin{tcolorbox}[]
Based on the detailed evaluation guideline and format requirement you provided, I'm now evaluating consistency of the following summary to the article: 
\\

```json\\
\{\\
  "article":"A woman \dots",\\
  "summary":"The mother \dots",\\
  "consistency\_score": \blue{0.7}\\
\}\\
```~~~~~~~~~~~~~~~~~~~(c)
\end{tcolorbox}
\end{minipage}
\begin{minipage}[c]{.24\textwidth}
\centering
\begin{tcolorbox}[]
Based on the detailed evaluation guideline and format requirement you provided, I'm now evaluating consistency of the following summary to the article:
\\

```json\\
\{\\
  "article":"A woman \dots",\\
  "summary":"The mother \dots",\\
  "consistency\_score": \\
\}\\
```~~~~~~~~~~~~~~~~~~~(d)
\end{tcolorbox}
\end{minipage}
\caption{Numerical sensitivity ablation examples from QAGS-X: (a) the original meta-prompt for evaluation prompt generation; (b) rounding evaluation score to one decimal place; (c) removing evaluation score range; (d) removing both score range and human evaluation scores.}
\label{fig:ablation_scores_in_prompts}
\end{figure*}

\noindent\textbf{Numerical Sensitivity.}~~The input to the inverse models for generating evaluation prompts includes two types of numerical information: the range of evaluation scores and the human score for the corresponding evaluation sample.
Additionally, human scores often contain multiple decimal places, as they are typically averaged across multiple human evaluators or, in the case of machine translation tasks using metrics like MQM, calculated by applying weighted aggregation across different scoring dimensions. 
This raises an interesting question: \textit{how sensitive are the inverse models to this numerical information, and what impact does it have on the quality and effectiveness of the generated evaluation prompts?} 
To investigate this, we conducted an ablation study by altering the numerical information in the original input: (i) rounding the human evaluation scores to one decimal place; (ii) removing the evaluation score range; and (iii) removing both the score range and the human evaluation score.
Examples of each input modification are shown in Figure~\ref{fig:ablation_scores_in_prompts}.

As shown in Table~\ref{tab:ablation_study}, all three ablation settings lead to only marginal decreases in model performance. Even in the worst case, where all numerical information is removed, the performance drops by only 5\% compared to using the original input with full information.
This suggests that, although the evaluation score range and ground-truth human evaluation scores might intuitively seem important, they have relatively minor impact on the quality of the generated evaluation prompts.
Examining the prompts generated from the ablation studies (see Figure~\ref{fig:ablation_study_prompts}) reveals that, in the \textit{w/o score range} setting, the inverse model is able to, in many cases,  infer the original evaluation score range based solely on the human evaluation scores. 
For example, given a score of 68 in the WMT evaluation task, the inverse model frequently generates a prompt template that adopts a 0 to 100 scale.
In the setting where no numerical information is provided, the model tends to randomly select a commonly used score range (e.g., 0–100 or 1–10), yet the resulting performance remains relatively stable across different score ranges.


\subsection{Model Size Scaling} 
\label{subsec:model_size_scaling}

\begin{table*}[h!]
    \centering
    \begin{adjustbox}{max width=0.95\textwidth}
    \begin{tabular}{lcccccccccccc}
        \toprule
        \textbf{\texttt{Evaluator}} & \multicolumn{2}{c}{\textbf{SummEval}} & \multicolumn{2}{c}{\textbf{QAGS-C}} & \multicolumn{2}{c}{\textbf{QAGS-X}} & \multicolumn{2}{c}{\textbf{Topical-Chat}} & \multicolumn{2}{c}{\textbf{WMT-22}} & \multicolumn{2}{c}{\textbf{Average}}\\
        \cmidrule(lr){2-3}\cmidrule(lr){4-5}\cmidrule(lr){6-7}\cmidrule(lr){8-9}\cmidrule(lr){10-11}\cmidrule(lr){12-13}
        \textit{~Evaluation Prompt} & $\rho$ & $r$ & $\rho$ & $r$ & $\rho$ & $r$ &  $\rho$ & $r$ & $\rho$ & $r$ & $\rho$ & $r$ \\
        \midrule
        \texttt{Qwen-2.5-\textbf{14B}-Instruct} \\
        \textit{~Human-Crafted Prompt} & \underline{0.450} & \underline{0.463} & \underline{0.687} & \underline{0.688} & \underline{0.539} & \underline{0.527} & \underline{0.587} & \underline{0.564} & \underline{0.299} & \underline{0.312} & \underline{0.512} & \underline{0.511}\\
        \textit{~Forward Prompt} & 0.417 & 0.443 & 0.612 & 0.653 & 0.261 & 0.264 & 0.568 & 0.560 & 0.291 & 0.301 & 0.430 & 0.444\\
        \textit{~Inverse Prompt (\textbf{Ours})} & \textbf{0.456} & \textbf{0.471} & \textbf{0.721} & \textbf{0.721} & \textbf{0.592} & \textbf{0.558} & \textbf{0.625} & \textbf{0.615} & \textbf{0.306} & \textbf{0.323} & \textbf{0.540} & \textbf{0.538}\\
        \midrule
        \texttt{Qwen-2.5-\textbf{7B}-Instruct} \\
        \textit{~Human-Crafted Prompt} & \underline{0.374} & \underline{0.430} & \underline{0.654} & \underline{0.668} & \underline{0.483} & \underline{0.464} & 0.398 & 0.393 & 0.271 & 0.202 & \underline{0.436} & \underline{0.431}\\
        \textit{~Forward Prompt} & 0.315 & 0.339 & 0.529 & 0.603 & 0.198 & 0.207 & \underline{0.436} & \underline{0.439} & \underline{0.274} & \underline{0.284} & 0.350 & 0.374\\
        \textit{~Inverse Prompt (\textbf{Ours})} & \textbf{0.418} & \textbf{0.457} & \textbf{0.661} & \textbf{0.673} & \textbf{0.524} & \textbf{0.530} & \textbf{0.502} & \textbf{0.501} & \textbf{0.313} & \textbf{0.316} & \textbf{0.484} & \textbf{0.495}\\
        \midrule
        \texttt{Qwen-2.5-\textbf{3B}-Instruct} \\
        \textit{~Human-Crafted Prompt} & \textbf{0.385} & \textbf{0.458} & \underline{0.516} & \underline{0.506} & \underline{0.438} & \underline{0.405} & \underline{0.338} & \underline{0.332} & 0.252 & 0.257 & \underline{0.386} & \underline{0.393}\\
        \textit{~Forward Prompt} & 0.243 & 0.253 & 0.427 & 0.409 & 0.129 & 0.118 & 0.311 & 0.291 & \underline{0.260} & \underline{0.309} & 0.274 & 0.276\\
        \textit{~Inverse Prompt (\textbf{Ours})} & \underline{0.339} & \underline{0.341} & \textbf{0.591} & \textbf{0.569} & \textbf{0.439} & \textbf{0.443} & \textbf{0.340} & \textbf{0.336} & \textbf{0.286} & \textbf{0.312} & \textbf{0.399} & \textbf{0.400}\\
        \bottomrule
    \end{tabular}
    \end{adjustbox}
    \caption{Results of the model scaling study.}
    \label{tab:model_size_result}
    \vspace{-3.5mm}
\end{table*}

To explore the impact of model size scaling on inversion learning, we trained inverse models with Qwen at multiple scales (3B, 7B, and 14B) in the black-box setting using the same Infinity-Instruct dataset comprising 660k samples. Table~\ref{tab:model_size_result} presents the results of these experiments, clearly demonstrating a positive correlation between model size and evaluation performance across all datasets and tasks. 
For example, the average Spearman correlation for inverse prompts increases from 0.399 with the 3B model to 0.540 with the 14B model, corresponding to 35\% of relative improvement, highlighting the effectiveness of model scaling.

Moreover, inverse prompts consistently outperform forward prompts and achieve higher correlations than human-crafted prompts across all model sizes, with the sole exception of the 3B model on the SummEval dataset.
These results validate the effectiveness of our inverse prompt generation method under model scaling.

\subsection{Case Study}

To analyse why forward, human-crafted, and inverse prompts exhibit different levels of effectiveness, we conducted a qualitative comparison of prompts generated based on the meta-prompt shown in Figure~\ref{fig:ablation_scores_in_prompts}(a), which contains the one-shot evaluation sample focused exclusively on the consistency dimension.  
Both the forward and inverse prompts were generated by \texttt{Qwen-2.5-7B-Instruct} under the black-box setting (cf. Table~\ref{tab:main_result}). The complete prompt examples are provided in Figure~\ref{fig:qwen_prompts} in Appendix~\ref{app_examples}.
For clarity, we highlight the sections corresponding to \darkgreen{Model Instruction} (e.g., role assignment), \darkred{Evaluation Criteria}, and \textcolor{blue}{Evaluation Guideline}.

Among the three types of prompts, the generated forward prompts define evaluation across multiple dimensions (e.g., comprehensiveness, accuracy, etc.), which does not align with the one-shot example used for evaluation prompt generation, where the focus is solely on assessing consistency.  
Comparing inversion and human-crafted prompts, there are several distinct differences in terms of their criteria descriptions, structure, and tone. For instance, inverse prompts explicitly assign a role to the model, framing it as ``\textit{an advanced AI assistant}'', which helps anchor the model’s perspective and behaviour during evaluation. In contrast, human-crafted prompts use a more natural and instructional tone without explicit role-playing, making them more approachable for human readers.

In terms of evaluation guidelines, inverse prompt provides step-by-step procedures with detailed and imperative phrasing (e.g. ``\textit{To perform the task, you must ...}''). Human-crafted prompt also includes task steps but present them more loosely, reflecting how humans naturally approach annotation tasks. For the consistency criterion, inverse prompts offer the most operational definition among the three, framing factual consistency through formal entailment-based reasoning. In comparison, human-crafted prompts describe it in simpler and less precise terms, emphasising factual alignment and penalising hallucinations. This formal, entailment-based framing in inverse prompts likely contributes to their effectiveness in evaluation tasks.
Additionally, inverse prompts use a continuous 0–1 scoring scale for fine-grained evaluation, whereas human-crafted prompts use 1–5 Likert scale.

Comparing the prompts generated by Qwen and LLaMA (see Figure~\ref{fig:llama_prompts}), we observe that the forward prompt from LLaMA is similar to Qwen’s but introduces even more evaluation criteria for irrelevant dimensions. For inverse prompts, LLaMA’s prompt is less formal (i.e., more conversational) and adopts a more instructional rather than assertive tone, offering intuitive but less rigorously defined descriptions of factual consistency, along with fewer procedural details.
Structurally, Qwen clearly separates model instruction, evaluation criteria, and evaluation guidelines, whereas LLaMA blends these elements more loosely.
This highlights that Qwen and LLaMA have different preferences in prompt style, which make them most effective.

In summary, our analysis supports the hypothesis that generating model-specific prompts is crucial, and that human-crafted prompts and guidelines do not necessarily translate into more effective prompts for LLMs.

\section{Conclusion}

In this work, we introduced a novel approach for generating high-quality, model-specific evaluation prompts through inversion learning, marking a sharp departure from practices that rely on human-crafted prompts. These hand-crafted prompts are often costly to produce and typically applied without considering their effectiveness across different LLMs. 
Our method eliminates the need for manual prompt engineering and outperforms both human-crafted prompts and strong prompt optimisation techniques, despite using only a single annotated example. In contrast, prompt optimisation methods typically require large validation sets and iterative tuning.
Extensive experiments on two open-source LLM families and a wide range of generation tasks demonstrate that our method can efficiently produce high-quality prompts from a single evaluation sample. Moreover, our results confirm the hypothesis that model-tailored prompts are essential for improving evaluation performance. Ultimately, this work contributes toward a new direction for more robust and efficient LLM-based evaluation.


\bibliography{custom}

\begin{thebibliography}{52}
\expandafter\ifx\csname natexlab\endcsname\relax\def\natexlab#1{#1}\fi

\bibitem[{Aher et~al.(2023)Aher, Arriaga, and Kalai}]{aher2023using}
Gati~V Aher, Rosa~I Arriaga, and Adam~Tauman Kalai. 2023.
\newblock Using large language models to simulate multiple humans and replicate human subject studies.
\newblock In \emph{International Conference on Machine Learning}, pages 337--371. PMLR.

\bibitem[{Belz et~al.(2024)Belz, Sedoc, Thomson, Mille, and Huidrom}]{belz2024inlg}
Anja Belz, Jo{\~a}o Sedoc, Craig Thomson, Simon Mille, and Rudali Huidrom. 2024.
\newblock The inlg 2024 tutorial on human evaluation of nlp system quality: Background, overall aims, and summaries of taught units.
\newblock In \emph{Proceedings of the 17th International Natural Language Generation Conference: Tutorial Abstract}, pages 1--12.

\bibitem[{Cao et~al.(2024)Cao, Cai, Zhang, Zou, and Lam}]{cao2024worstpromptperformancelarge}
Bowen Cao, Deng Cai, Zhisong Zhang, Yuexian Zou, and Wai Lam. 2024.
\newblock \href {http://arxiv.org/abs/2406.10248} {On the worst prompt performance of large language models}.

\bibitem[{Carlini et~al.(2021)Carlini, Tram{\`e}r, Wallace, Jagielski, Herbert-Voss, Lee, Roberts, Brown, Song, Erlingsson, Oprea, and Raffel}]{Nicholas2021Extracting}
Nicholas Carlini, Florian Tram{\`e}r, Eric Wallace, Matthew Jagielski, Ariel Herbert-Voss, Katherine Lee, Adam Roberts, Tom Brown, Dawn Song, {\'U}lfar Erlingsson, Alina Oprea, and Colin Raffel. 2021.
\newblock \href {https://www.usenix.org/conference/usenixsecurity21/presentation/carlini-extracting} {Extracting training data from large language models}.
\newblock In \emph{30th USENIX Security Symposium (USENIX Security 21)}, pages 2633--2650. USENIX Association.

\bibitem[{Elangovan et~al.(2024)Elangovan, Liu, Xu, Bodapati, and Roth}]{elangovan-etal-2024-considers}
Aparna Elangovan, Ling Liu, Lei Xu, Sravan~Babu Bodapati, and Dan Roth. 2024.
\newblock \href {https://doi.org/10.18653/v1/2024.acl-long.63} {{C}on{S}i{DERS}-the-human evaluation framework: Rethinking human evaluation for generative large language models}.
\newblock In \emph{Proceedings of the 62nd Annual Meeting of the Association for Computational Linguistics (Volume 1: Long Papers)}, pages 1137--1160, Bangkok, Thailand. Association for Computational Linguistics.

\bibitem[{Errica et~al.(2024)Errica, Siracusano, Sanvito, and Bifulco}]{errica2024did}
Federico Errica, Giuseppe Siracusano, Davide Sanvito, and Roberto Bifulco. 2024.
\newblock What did i do wrong? quantifying llms' sensitivity and consistency to prompt engineering.
\newblock \emph{arXiv preprint arXiv:2406.12334}.

\bibitem[{Fabbri et~al.(2021)Fabbri, Kry{\'s}ci{\'n}ski, McCann, Xiong, Socher, and Radev}]{fabbri-etal-2021-summeval}
Alexander~R. Fabbri, Wojciech Kry{\'s}ci{\'n}ski, Bryan McCann, Caiming Xiong, Richard Socher, and Dragomir Radev. 2021.
\newblock \href {https://doi.org/10.1162/tacl_a_00373} {{S}umm{E}val: Re-evaluating summarization evaluation}.
\newblock \emph{Transactions of the Association for Computational Linguistics}, 9:391--409.

\bibitem[{Freitag et~al.(2022)Freitag, Rei, Mathur, Lo, Stewart, Avramidis, Kocmi, Foster, Lavie, and Martins}]{freitag-etal-2022-results}
Markus Freitag, Ricardo Rei, Nitika Mathur, Chi-kiu Lo, Craig Stewart, Eleftherios Avramidis, Tom Kocmi, George Foster, Alon Lavie, and Andr{\'e} F.~T. Martins. 2022.
\newblock \href {https://aclanthology.org/2022.wmt-1.2/} {Results of {WMT}22 metrics shared task: Stop using {BLEU} {--} neural metrics are better and more robust}.
\newblock In \emph{Proceedings of the Seventh Conference on Machine Translation (WMT)}, pages 46--68, Abu Dhabi, United Arab Emirates (Hybrid). Association for Computational Linguistics.

\bibitem[{Gao et~al.(2025)Gao, Hu, Lin, and Wan}]{gao2025analyzingevaluatingcorrelationmeasures}
Mingqi Gao, Xinyu Hu, Li~Lin, and Xiaojun Wan. 2025.
\newblock \href {http://arxiv.org/abs/2410.16834} {Analyzing and evaluating correlation measures in nlg meta-evaluation}.

\bibitem[{Gopalakrishnan et~al.(2019)Gopalakrishnan, Hedayatnia, Chen, Gottardi, Kwatra, Venkatesh, Gabriel, and Hakkani-T{\"u}r}]{Gopalakrishnan2019TopicalChatTK}
Karthik Gopalakrishnan, Behnam Hedayatnia, Qinlang Chen, Anna Gottardi, Sanjeev Kwatra, Anu Venkatesh, Raefer Gabriel, and Dilek~Z. Hakkani-T{\"u}r. 2019.
\newblock \href {https://api.semanticscholar.org/CorpusID:202717047} {Topical-chat: Towards knowledge-grounded open-domain conversations}.
\newblock \emph{ArXiv}, abs/2308.11995.

\bibitem[{Grattafiori et~al.(2024)Grattafiori, Dubey, Jauhri, Pandey, Kadian, Al-Dahle, Letman, Mathur, Schelten, Vaughan, and et~al.}]{grattafiori2024llama3herdmodels}
Aaron Grattafiori, Abhimanyu Dubey, Abhinav Jauhri, Abhinav Pandey, Abhishek Kadian, Ahmad Al-Dahle, Aiesha Letman, Akhil Mathur, Alan Schelten, Alex Vaughan, and Amy~Yang et~al. 2024.
\newblock \href {http://arxiv.org/abs/2407.21783} {The llama 3 herd of models}.

\bibitem[{Holtzman et~al.(2020)Holtzman, Buys, Du, Forbes, and Choi}]{holtzman2020curiouscaseneuraltext}
Ari Holtzman, Jan Buys, Li~Du, Maxwell Forbes, and Yejin Choi. 2020.
\newblock \href {http://arxiv.org/abs/1904.09751} {The curious case of neural text degeneration}.

\bibitem[{Howcroft et~al.(2020)Howcroft, Belz, Clinciu, Gkatzia, Hasan, Mahamood, Mille, van Miltenburg, Santhanam, and Rieser}]{howcroft-etal-2020-twenty}
David~M. Howcroft, Anya Belz, Miruna-Adriana Clinciu, Dimitra Gkatzia, Sadid~A. Hasan, Saad Mahamood, Simon Mille, Emiel van Miltenburg, Sashank Santhanam, and Verena Rieser. 2020.
\newblock \href {https://doi.org/10.18653/v1/2020.inlg-1.23} {Twenty years of confusion in human evaluation: {NLG} needs evaluation sheets and standardised definitions}.
\newblock In \emph{Proceedings of the 13th International Conference on Natural Language Generation}, pages 169--182.

\bibitem[{Hu et~al.(2022)Hu, yelong shen, Wallis, Allen-Zhu, Li, Wang, Wang, and Chen}]{hu2022lora}
Edward~J Hu, yelong shen, Phillip Wallis, Zeyuan Allen-Zhu, Yuanzhi Li, Shean Wang, Lu~Wang, and Weizhu Chen. 2022.
\newblock \href {https://openreview.net/forum?id=nZeVKeeFYf9} {Lo{RA}: Low-rank adaptation of large language models}.
\newblock In \emph{International Conference on Learning Representations}.

\bibitem[{Huijzer and Hill(2023)}]{huijzer2023llm}
Rik Huijzer and Yannick Hill. 2023.
\newblock \href {https://doi.org/10.31234/osf.io/munc9} {Large language models show human behavior}.
\newblock Workingpaper, PsyArXiv.

\bibitem[{Kocmi and Federmann(2023)}]{kocmi-federmann-2023-large}
Tom Kocmi and Christian Federmann. 2023.
\newblock \href {https://aclanthology.org/2023.eamt-1.19/} {Large language models are state-of-the-art evaluators of translation quality}.
\newblock In \emph{Proceedings of the 24th Annual Conference of the European Association for Machine Translation}, pages 193--203, Tampere, Finland. European Association for Machine Translation.

\bibitem[{Lee et~al.(2025)Lee, Lim, Han, Oh, Chae, Chung, Kim, Kwak, Lee, Lee, Yeo, and Yu}]{lee-etal-2025-llms}
Seungbeen Lee, Seungwon Lim, Seungju Han, Giyeong Oh, Hyungjoo Chae, Jiwan Chung, Minju Kim, Beong-woo Kwak, Yeonsoo Lee, Dongha Lee, Jinyoung Yeo, and Youngjae Yu. 2025.
\newblock \href {https://doi.org/10.18653/v1/2025.findings-naacl.469} {Do {LLM}s have distinct and consistent personality? {TRAIT}: Personality testset designed for {LLM}s with psychometrics}.
\newblock In \emph{Findings of the Association for Computational Linguistics: NAACL 2025}, pages 8397--8437, Albuquerque, New Mexico. Association for Computational Linguistics.

\bibitem[{Leiter and Eger(2024)}]{leiter-eger-2024-prexme}
Christoph Leiter and Steffen Eger. 2024.
\newblock \href {https://doi.org/10.18653/v1/2024.emnlp-main.641} {{P}r{E}x{M}e! large scale prompt exploration of open source {LLM}s for machine translation and summarization evaluation}.
\newblock In \emph{Proceedings of the 2024 Conference on Empirical Methods in Natural Language Processing}, pages 11481--11506, Miami, Florida, USA. Association for Computational Linguistics.

\bibitem[{Li et~al.(2024)Li, Dong, Chen, Su, Zhou, Ai, Ye, and Liu}]{li2024llmsasjudgescomprehensivesurveyllmbased}
Haitao Li, Qian Dong, Junjie Chen, Huixue Su, Yujia Zhou, Qingyao Ai, Ziyi Ye, and Yiqun Liu. 2024.
\newblock \href {http://arxiv.org/abs/2412.05579} {Llms-as-judges: A comprehensive survey on llm-based evaluation methods}.

\bibitem[{Li et~al.(2025)Li, Du, Zhao, wen Zhang, Wang, Gao, Liu, and Lin}]{li2025infinityinstructscalinginstruction}
Jijie Li, Li~Du, Hanyu Zhao, Bo~wen Zhang, Liangdong Wang, Boyan Gao, Guang Liu, and Yonghua Lin. 2025.
\newblock \href {http://arxiv.org/abs/2506.11116} {Infinity instruct: Scaling instruction selection and synthesis to enhance language models}.

\bibitem[{Liu et~al.(2023)Liu, Iter, Xu, Wang, Xu, and Zhu}]{liu-etal-2023-g}
Yang Liu, Dan Iter, Yichong Xu, Shuohang Wang, Ruochen Xu, and Chenguang Zhu. 2023.
\newblock \href {https://doi.org/10.18653/v1/2023.emnlp-main.153} {{G}-eval: {NLG} evaluation using gpt-4 with better human alignment}.
\newblock In \emph{Proceedings of the 2023 Conference on Empirical Methods in Natural Language Processing}, pages 2511--2522, Singapore. Association for Computational Linguistics.

\bibitem[{Liu et~al.(2024)Liu, Moosavi, and Lin}]{liu-etal-2024-llms-narcissistic}
Yiqi Liu, Nafise Moosavi, and Chenghua Lin. 2024.
\newblock \href {https://doi.org/10.18653/v1/2024.findings-acl.753} {{LLM}s as narcissistic evaluators: When ego inflates evaluation scores}.
\newblock In \emph{Findings of the Association for Computational Linguistics: ACL 2024}, pages 12688--12701, Bangkok, Thailand. Association for Computational Linguistics.

\bibitem[{Loakman et~al.(2023)Loakman, Maladry, and Lin}]{loakman-etal-2023-iron}
Tyler Loakman, Aaron Maladry, and Chenghua Lin. 2023.
\newblock \href {https://doi.org/10.18653/v1/2023.findings-emnlp.444} {The iron(ic) melting pot: Reviewing human evaluation in humour, irony and sarcasm generation}.
\newblock In \emph{Findings of the Association for Computational Linguistics: EMNLP 2023}, pages 6676--6689.

\bibitem[{Morris et~al.(2023{\natexlab{a}})Morris, Kuleshov, Shmatikov, and Rush}]{morris2023text}
John~X Morris, Volodymyr Kuleshov, Vitaly Shmatikov, and Alexander~M Rush. 2023{\natexlab{a}}.
\newblock Text embeddings reveal (almost) as much as text.
\newblock \emph{arXiv preprint arXiv:2310.06816}.

\bibitem[{Morris et~al.(2023{\natexlab{b}})Morris, Zhao, Chiu, Shmatikov, and Rush}]{morris2023languagemodelinversion}
John~X. Morris, Wenting Zhao, Justin~T. Chiu, Vitaly Shmatikov, and Alexander~M. Rush. 2023{\natexlab{b}}.
\newblock \href {http://arxiv.org/abs/2311.13647} {Language model inversion}.

\bibitem[{Muñoz-Ortiz et~al.(2024)Muñoz-Ortiz, Gómez-Rodríguez, and Vilares}]{Mu_oz_Ortiz_2024}
Alberto Muñoz-Ortiz, Carlos Gómez-Rodríguez, and David Vilares. 2024.
\newblock \href {https://doi.org/10.1007/s10462-024-10903-2} {Contrasting linguistic patterns in human and llm-generated news text}.
\newblock \emph{Artificial Intelligence Review}, 57(10).

\bibitem[{Petrov et~al.(2024)Petrov, Dimitrov, Baader, Müller, and Vechev}]{petrov2024dagerexactgradientinversion}
Ivo Petrov, Dimitar~I. Dimitrov, Maximilian Baader, Mark~Niklas Müller, and Martin Vechev. 2024.
\newblock \href {http://arxiv.org/abs/2405.15586} {Dager: Exact gradient inversion for large language models}.

\bibitem[{Polo et~al.(2024)Polo, Xu, Weber, Silva, Bhardwaj, Choshen, de~Oliveira, Sun, and Yurochkin}]{Polo2024EfficientME}
Felipe~Maia Polo, Ronald Xu, Lucas Weber, M'irian Silva, Onkar Bhardwaj, Leshem Choshen, Allysson Flavio~Melo de~Oliveira, Yuekai Sun, and Mikhail Yurochkin. 2024.
\newblock \href {https://api.semanticscholar.org/CorpusID:270062938} {Efficient multi-prompt evaluation of llms}.
\newblock \emph{ArXiv}, abs/2405.17202.

\bibitem[{Pryzant et~al.(2023)Pryzant, Iter, Li, Lee, Zhu, and Zeng}]{pryzant-etal-2023-automatic}
Reid Pryzant, Dan Iter, Jerry Li, Yin Lee, Chenguang Zhu, and Michael Zeng. 2023.
\newblock \href {https://doi.org/10.18653/v1/2023.emnlp-main.494} {Automatic prompt optimization with ``gradient descent'' and beam search}.
\newblock In \emph{Proceedings of the 2023 Conference on Empirical Methods in Natural Language Processing}, pages 7957--7968, Singapore. Association for Computational Linguistics.

\bibitem[{Qian et~al.(2024)Qian, Sindhujan, Kabra, Kanojia, Orasan, Ranasinghe, and Blain}]{qian-etal-2024-large}
Shenbin Qian, Archchana Sindhujan, Minnie Kabra, Diptesh Kanojia, Constantin Orasan, Tharindu Ranasinghe, and Fred Blain. 2024.
\newblock \href {https://doi.org/10.18653/v1/2024.emnlp-main.214} {What do large language models need for machine translation evaluation?}
\newblock In \emph{Proceedings of the 2024 Conference on Empirical Methods in Natural Language Processing}, pages 3660--3674, Miami, Florida, USA. Association for Computational Linguistics.

\bibitem[{{Qwen Team}(2024)}]{qwen2.5}
{Qwen Team}. 2024.
\newblock \href {https://qwenlm.github.io/blog/qwen2.5/} {Qwen2.5: A party of foundation models}.

\bibitem[{Rafailov et~al.(2023)Rafailov, Sharma, Mitchell, Manning, Ermon, and Finn}]{rafailov2023direct}
Rafael Rafailov, Archit Sharma, Eric Mitchell, Christopher~D Manning, Stefano Ermon, and Chelsea Finn. 2023.
\newblock \href {https://openreview.net/forum?id=HPuSIXJaa9} {Direct preference optimization: Your language model is secretly a reward model}.
\newblock In \emph{Thirty-seventh Conference on Neural Information Processing Systems}.

\bibitem[{Ramnath et~al.(2025)Ramnath, Zhou, Guan, Mishra, Qi, Shen, Wang, Woo, Jeoung, Wang, Wang, Ding, Lu, Xu, Zhou, Srinivasan, Yan, Chen, Ding, Xu, and Cheong}]{ramnath-etal-2025-apo-survey}
Kiran Ramnath, Kang Zhou, Sheng Guan, Soumya~Smruti Mishra, Xuan Qi, Zhengyuan Shen, Shuai Wang, Sangmin Woo, Sullam Jeoung, Yawei Wang, Haozhu Wang, Han Ding, Yuzhe Lu, Zhichao Xu, Yun Zhou, Balasubramaniam Srinivasan, Qiaojing Yan, Yueyan Chen, Haibo Ding, Panpan Xu, and Lin~Lee Cheong. 2025.
\newblock A systematic survey of automatic prompt optimization techniques.
\newblock \emph{arXiv preprint arXiv:2502.16923}.

\bibitem[{Sclar et~al.(2024)Sclar, Choi, Tsvetkov, and Suhr}]{sclar2024quantifyinglanguagemodelssensitivity}
Melanie Sclar, Yejin Choi, Yulia Tsvetkov, and Alane Suhr. 2024.
\newblock \href {http://arxiv.org/abs/2310.11324} {Quantifying language models' sensitivity to spurious features in prompt design or: How i learned to start worrying about prompt formatting}.

\bibitem[{Shin et~al.(2020)Shin, Razeghi, Logan~IV, Wallace, and Singh}]{shin-etal-2020-autoprompt}
Taylor Shin, Yasaman Razeghi, Robert~L. Logan~IV, Eric Wallace, and Sameer Singh. 2020.
\newblock \href {https://doi.org/10.18653/v1/2020.emnlp-main.346} {{A}uto{P}rompt: {E}liciting {K}nowledge from {L}anguage {M}odels with {A}utomatically {G}enerated {P}rompts}.
\newblock In \emph{Proceedings of the 2020 Conference on Empirical Methods in Natural Language Processing (EMNLP)}, pages 4222--4235, Online. Association for Computational Linguistics.

\bibitem[{Song and Raghunathan(2020)}]{song2020informationleakageembeddingmodels}
Congzheng Song and Ananth Raghunathan. 2020.
\newblock \href {http://arxiv.org/abs/2004.00053} {Information leakage in embedding models}.

\bibitem[{Sun et~al.(2023)Sun, Li, Xu, Homma, Cao, Wu, Jiao, and Charles}]{sun2023autohint}
Hong Sun, Xue Li, Yinchuan Xu, Youkow Homma, Qi~Cao, Min Wu, Jian Jiao, and Denis Charles. 2023.
\newblock Autohint: Automatic prompt optimization with hint generation.
\newblock \emph{arXiv preprint arXiv:2307.07415}.

\bibitem[{Sun et~al.(2025)Sun, Yin, Xu, Kolter, and Liu}]{sun2025idiosyncrasies}
Mingjie Sun, Yida Yin, Zhiqiu Xu, J~Zico Kolter, and Zhuang Liu. 2025.
\newblock \href {https://openreview.net/forum?id=FCZ3jVzmTZ} {Idiosyncrasies in large language models}.
\newblock In \emph{Forty-second International Conference on Machine Learning}.

\bibitem[{Wang et~al.(2020)Wang, Cho, and Lewis}]{wang-etal-2020-asking}
Alex Wang, Kyunghyun Cho, and Mike Lewis. 2020.
\newblock \href {https://doi.org/10.18653/v1/2020.acl-main.450} {Asking and answering questions to evaluate the factual consistency of summaries}.
\newblock In \emph{Proceedings of the 58th Annual Meeting of the Association for Computational Linguistics}, pages 5008--5020, Online. Association for Computational Linguistics.

\bibitem[{Wang et~al.(2023)Wang, Liang, Meng, Sun, Shi, Li, Xu, Qu, and Zhou}]{wang-etal-2023-chatgpt}
Jiaan Wang, Yunlong Liang, Fandong Meng, Zengkui Sun, Haoxiang Shi, Zhixu Li, Jinan Xu, Jianfeng Qu, and Jie Zhou. 2023.
\newblock \href {https://doi.org/10.18653/v1/2023.newsum-1.1} {Is {C}hat{GPT} a good {NLG} evaluator? a preliminary study}.
\newblock In \emph{Proceedings of the 4th New Frontiers in Summarization Workshop}, pages 1--11, Singapore. Association for Computational Linguistics.

\bibitem[{Wen et~al.(2025)Wen, Ke, Sun, Wang, Gu, Zhou, Tang, Wang, and Huang}]{wen-etal-2025-hpss}
Bosi Wen, Pei Ke, Yufei Sun, Cunxiang Wang, Xiaotao Gu, Jinfeng Zhou, Jie Tang, Hongning Wang, and Minlie Huang. 2025.
\newblock \href {https://doi.org/10.18653/v1/2025.findings-acl.1282} {{HPSS}: Heuristic prompting strategy search for {LLM} evaluators}.
\newblock In \emph{Findings of the Association for Computational Linguistics: ACL 2025}, pages 24974--25007, Vienna, Austria. Association for Computational Linguistics.

\bibitem[{Xie et~al.(2025)Xie, Cai, Wang, Li, Sang, Yang, Zhang, Li, Zhu, Liu et~al.}]{xie2025infir}
Congkai Xie, Shuo Cai, Wenjun Wang, Pengxiang Li, Zhijie Sang, Kejing Yang, Yiming Zhang, Zhen Li, Guanghao Zhu, Zeyu Liu, et~al. 2025.
\newblock Infir: Crafting effective small language models and multimodal small language models in reasoning.
\newblock \emph{arXiv preprint arXiv:2502.11573}.

\bibitem[{Yang et~al.(2023)Yang, Wang, Lu, Liu, Le, Zhou, and Chen}]{yang-etal-2023-opro}
Chengrun Yang, Xuezhi Wang, Yifeng Lu, Hanxiao Liu, Quoc~V. Le, Denny Zhou, and Xinyun Chen. 2023.
\newblock Large language models as optimizers.
\newblock \emph{arXiv preprint arXiv:2309.03409}.
\newblock ICLR 2024.

\bibitem[{Yuan et~al.(2021)Yuan, Neubig, and Liu}]{NEURIPS2021_e4d2b6e6}
Weizhe Yuan, Graham Neubig, and Pengfei Liu. 2021.
\newblock \href {https://proceedings.neurips.cc/paper/2021/file/e4d2b6e6fdeca3e60e0f1a62fee3d9dd-Paper.pdf} {Bartscore: Evaluating generated text as text generation}.
\newblock In \emph{Advances in Neural Information Processing Systems}, volume~34, pages 27263--27277. Curran Associates, Inc.

\bibitem[{Zhang et~al.(2024)Zhang, Morris, and Shmatikov}]{zhang2024extractingpromptsinvertingllm}
Collin Zhang, John~X. Morris, and Vitaly Shmatikov. 2024.
\newblock \href {http://arxiv.org/abs/2405.15012} {Extracting prompts by inverting llm outputs}.

\bibitem[{Zhang et~al.(2020)Zhang, Kishore, Wu, Weinberger, and Artzi}]{Zhang2020BERTScore}
Tianyi Zhang, Varsha Kishore, Felix Wu, Kilian~Q. Weinberger, and Yoav Artzi. 2020.
\newblock \href {https://openreview.net/forum?id=SkeHuCVFDr} {Bertscore: Evaluating text generation with bert}.
\newblock In \emph{International Conference on Learning Representations}.

\bibitem[{Zhao et~al.(2023)Zhao, Yang, Lin, Rong, Villavicencio, and Cui}]{zhao-etal-2023-evaluating}
Kun Zhao, Bohao Yang, Chenghua Lin, Wenge Rong, Aline Villavicencio, and Xiaohui Cui. 2023.
\newblock Evaluating open-domain dialogues in latent space with next sentence prediction and mutual information.
\newblock In \emph{Proceedings of the 61st Annual Meeting of the Association for Computational Linguistics (Volume 1: Long Papers)}.

\bibitem[{Zhao et~al.(2024)Zhao, Yang, Tang, Lin, and Zhan}]{zhao-etal-2024-slide}
Kun Zhao, Bohao Yang, Chen Tang, Chenghua Lin, and Liang Zhan. 2024.
\newblock {SLIDE}: A framework integrating small and large language models for open-domain dialogues evaluation.
\newblock In \emph{Findings of the Association for Computational Linguistics: ACL 2024}.

\bibitem[{Zhao et~al.(2025)Zhao, Zhou, Li, Tang, Wang, Hou, Min, Zhang, Zhang, Dong, Du, Yang, Chen, Chen, Jiang, Ren, Li, Tang, Liu, Liu, Nie, and Wen}]{zhao2025surveylargelanguagemodels}
Wayne~Xin Zhao, Kun Zhou, Junyi Li, Tianyi Tang, Xiaolei Wang, Yupeng Hou, Yingqian Min, Beichen Zhang, Junjie Zhang, Zican Dong, Yifan Du, Chen Yang, Yushuo Chen, Zhipeng Chen, Jinhao Jiang, Ruiyang Ren, Yifan Li, Xinyu Tang, Zikang Liu, Peiyu Liu, Jian-Yun Nie, and Ji-Rong Wen. 2025.
\newblock \href {http://arxiv.org/abs/2303.18223} {A survey of large language models}.

\bibitem[{Zheng et~al.(2023)Zheng, Chiang, Sheng, Zhuang, Wu, Zhuang, Lin, Li, Li, Xing, Zhang, Gonzalez, and Stoica}]{zheng2023judging}
Lianmin Zheng, Wei-Lin Chiang, Ying Sheng, Siyuan Zhuang, Zhanghao Wu, Yonghao Zhuang, Zi~Lin, Zhuohan Li, Dacheng Li, Eric Xing, Hao Zhang, Joseph~E. Gonzalez, and Ion Stoica. 2023.
\newblock \href {https://openreview.net/forum?id=uccHPGDlao} {Judging {LLM}-as-a-judge with {MT}-bench and chatbot arena}.
\newblock In \emph{Thirty-seventh Conference on Neural Information Processing Systems Datasets and Benchmarks Track}.

\bibitem[{Zheng et~al.(2024)Zheng, Zhang, Zhang, Ye, Luo, Feng, and Ma}]{zheng2024llamafactory}
Yaowei Zheng, Richong Zhang, Junhao Zhang, Yanhan Ye, Zheyan Luo, Zhangchi Feng, and Yongqiang Ma. 2024.
\newblock \href {http://arxiv.org/abs/2403.13372} {Llamafactory: Unified efficient fine-tuning of 100+ language models}.
\newblock In \emph{Proceedings of the 62nd Annual Meeting of the Association for Computational Linguistics (Volume 3: System Demonstrations)}, Bangkok, Thailand. Association for Computational Linguistics.

\bibitem[{Zhong et~al.(2022)Zhong, Liu, Yin, Mao, Jiao, Liu, Zhu, Ji, and Han}]{zhong2022unifiedmultidimensionalevaluatortext}
Ming Zhong, Yang Liu, Da~Yin, Yuning Mao, Yizhu Jiao, Pengfei Liu, Chenguang Zhu, Heng Ji, and Jiawei Han. 2022.
\newblock \href {http://arxiv.org/abs/2210.07197} {Towards a unified multi-dimensional evaluator for text generation}.

\end{thebibliography}
\bibliographystyle{acl_natbib}

\clearpage

\appendix
\onecolumn
\section{Prompt Examples} 
\label{app_examples}
\begin{figure}[!hbt]
\centering
\tcbset{
    colback=gray!5!white,
    fontupper=\scriptsize,
    left=1pt,
    right=1pt,
    valign=center,
    height=.7\textheight
}
\begin{minipage}[c]{.32\textwidth}
\centering
\begin{tcolorbox}[]
{\color{red!75!black}To evaluate the consistency of a summary with the article, follow these criteria:

1. **Comprehensive Coverage**: Ensure the summary captures the main points and key details from the article without omitting crucial information. \\
2. **Accuracy**: Verify that the summary accurately reflects the content of the article, maintaining the correct facts and figures. \\
3. **Relevance**: Confirm that the summary is relevant to the article's topic and does not include irrelevant information. \\
4. **Precision**: Check that the summary uses precise language that aligns with the article's tone and style. \\
5. **Brevity**: Ensure the summary is concise and does not include unnecessary details or elongated phrases that detract from its clarity.}

{\color{blue}By adhering to these criteria, summaries will be evaluated for their consistency with the original article. \\}
Now please evaluate the following summary to the article based on the above guideline criteria:

Article: \\
\{article\}

Summary: \\
\{summary\}

Please just directly output the final consistency score in a json format.
For example: \\
```json \\
\{ \\
    "article": \{article\}, \\
    "summary": \{summary\}, \\
    "consistency\_score": <a score between 0 and 1> \\
\} \\
```
\end{tcolorbox}
\subcaption{Forward Prompt}
\end{minipage}
\begin{minipage}[c]{.32\textwidth}
\centering
\begin{tcolorbox}[]
{\color{green!50!black}You will be given a news article. You will then be given one summary written for this article.

Your task is to rate the summary on one metric.

Please make sure you read and understand these instructions carefully. Please keep this document open while reviewing, and refer to it as needed.}

{\color{red!75!black}Evaluation Criteria:

Consistency (1-5) - the factual alignment between the summary and the summarised source. A factually consistent summary contains only statements that are entailed by the source document. Annotators were also asked to penalise summaries that contained hallucinated facts.}

{\color{blue}Evaluation Steps:

1. Read the news article carefully and identify the main facts and details it presents. \\
2. Read the summary and compare it to the article. Check if the summary contains any factual errors that are not supported by the article. \\
3. Assign a score for consistency based on the Evaluation Criteria.}

Please evaluate the following summary:\\
Source Text: 

\{article\}

Summary: 

\{summary\}

Please just directly output the consistency score in a json format.\\
For example:\\
```json \\
\{\\
  "article": "content of the article", \\
  "summary": "content of the summary", \\
    "consistency\_score": <a score between 1 and 5> \\
\}\\
```
\end{tcolorbox}
\subcaption{Human-Crafted Prompt}
\end{minipage}
\begin{minipage}[c]{.32\textwidth}
\centering
\begin{tcolorbox}[]
{\color{green!50!black}You are an advanced AI assistant tasked with evaluating factual consistency of summaries based on detailed articles. Your role involves a thorough analysis of the article provided to ensure the summary aligns perfectly with the content and events described in the article. The summary should not contain any information that contradicts, misrepresents, or distorts the facts presented in the article.}

{\color{blue}To perform this task, you must:

1. Examine each sentence in the summary in relation to the article's content.\\
2. Identify any factual inconsistencies, such as misrepresentations, contradictions, or omitted key details.\\
3. Assign a factual consistency score to the summary on a scale of 0 to 1, where 1 indicates perfect factual consistency and 0 indicates complete factual inconsistency.\\}

{\color{red!75!black}For your reference, here is a detailed evaluation guideline and format requirement:

**Evaluation Guideline:**\\
- A summary is factually consistent if every sentence in the summary is logically entailed by the article and no contradictions are present.}

**Output Format:**\\
```json\\
\{ \\
  "article": "content of the article", \\
  "summary": "content of the summary", \\
  "consistency\_score": score between 0 and 1 \\
\} \\
```

By adhering to this evaluation guideline and format, you will ensure that the factual consistency of summaries is rigorously assessed. Please proceed to evaluate the summaries based on the articles provided.

Article:
\{article\}

Summary:
\{summary\}
\end{tcolorbox}
\subcaption{Inverse Prompt}
\end{minipage}
\caption{Prompts for Qwen on QAGS dataset. Green texts denote the \darkgreen{Model Instruction} part, red texts denote the \darkred{Evaluation Criteria} part,  and blue texts denote the \textcolor{blue}{Evaluation Procedure} part. The remaining part is the Input and Format Requirement.}
\label{fig:qwen_prompts}
\end{figure}

\begin{figure}[!hbt]
\centering
\tcbset{
    colback=gray!5!white,
    fontupper=\scriptsize,
    left=1pt,
    right=1pt,
    valign=center,
    height=.875\textheight
}
\begin{minipage}[c]{.32\textwidth}
\centering
\begin{tcolorbox}[]
{\color{red!75!black}Here is the evaluation guideline to assess the consistency of summaries:

**Consistency Evaluation Criteria**

1. **Factual Accuracy**: The summary accurately reflects the main facts and events described in the article.\\
2. **Main Idea Representation**: The summary effectively conveys the main idea or key takeaway from the article.\\
3. **Relevance**: The summary only includes information that is relevant to the article's main topic and events.\\
4. **Omission of Irrelevant Details**: The summary does not include unnecessary or tangential details that are not essential to understanding the article's main point.\\
5. **Cohesion**: The summary is logically organized and easy to follow, with a clear structure and flow of ideas.\\
6. **Language Clarity**: The summary uses clear and concise language, avoiding ambiguity and jargon.\\
7. **Accuracy of Quotes and Statements**: If the summary includes quotes or statements from individuals, they are accurately represented and not taken out of context.\\
8. **Comprehensive Coverage**: The summary provides a comprehensive overview of the article's main points, without omitting important details.\\
9. **Avoidance of Misrepresentation**: The summary does not misrepresent or distort the information presented in the article.\\
10. **Contextual Understanding**: The summary demonstrates an understanding of the article's context, including any relevant background information or relationships between events.}

{\color{blue}**Scoring**

By following these criteria, models can systematically assess the consistency of summaries with their respective articles, ensuring that evaluations are fair and comprehensive.\\}
Now please evaluate the following summary to the article based on the above guideline criteria:

Article:\\
\{article\}

Summary:
\{summary\}

Please just directly output the consistency score in a json format.\\
For example:\\
```json\\
\{\\
    "article": \{article\},\\
    "summary": \{summary\},\\
    "consistency\_score": <a score between 0 and 1>\\
\}\\
```
\end{tcolorbox}
\subcaption{Forward Prompt}
\end{minipage}
\begin{minipage}[c]{.32\textwidth}
\centering
\begin{tcolorbox}[]
{\color{green!50!black}You will be given a news article. You will then be given one summary written for this article.

Your task is to rate the summary on one metric.

Please make sure you read and understand these instructions carefully. Please keep this document open while reviewing, and refer to it as needed.}

{\color{red!75!black}Evaluation Criteria:

Consistency (1-5) - the factual alignment between the summary and the summarised source. A factually consistent summary contains only statements that are entailed by the source document. Annotators were also asked to penalise summaries that contained hallucinated facts.}

{\color{blue}Evaluation Steps:

1. Read the news article carefully and identify the main facts and details it presents. \\
2. Read the summary and compare it to the article. Check if the summary contains any factual errors that are not supported by the article. \\
3. Assign a score for consistency based on the Evaluation Criteria.}

Please evaluate the following summary:\\
Source Text: 

\{article\}

Summary: 

\{summary\}

Please just directly output the consistency score in a json format.\\
For example:\\
```json \\
\{\\
  "article": "content of the article", \\
  "summary": "content of the summary", \\
    "consistency\_score": <a score between 1 and 5> \\
\}\\
```
\end{tcolorbox}
\subcaption{Human-Crafted Prompt}
\end{minipage}
\begin{minipage}[c]{.32\textwidth}
\centering
\begin{tcolorbox}[]
{\color{green!50!black}You are an expert in evaluating the factual consistency of outputs generated by automatic summarization models (such as BART, T5, etc.). These models are known to sometimes produce summaries that are not factually consistent with the news articles they summarize.}
{\color{red!75!black}For example, the models sometimes hallucinate unmentioned facts, misrepresent the facts in the article, or introduce other factual mistakes. A summary is factually consistent if all the facts it presents are also explicitly mentioned in the news article.}

{\color{blue}In addition to the article and summary below, I would like you to assess the consistency of the summary with the article using a score between 0 and 1, with 1 indicating full consistency and 0 indicating no consistency.}

Here is the detailed evaluation guideline and format requirement:

1. Article: [article]\\
2. Summary: [summary]\\
3. Consistency Score: [consistency score]

```json\\
\{\\
  "article": \{article\},\\
  "summary": \{summary\},\\
  "consistency\_score": <a score between 0 and 1>\\
\}
```

'''
\end{tcolorbox}
\subcaption{Inverse Prompt}
\end{minipage}
\caption{Prompts for LLaMA on QAGS dataset.}
\label{fig:llama_prompts}
\end{figure}

\begin{figure}
\centering
\tcbset{
    colback=gray!5!white,
    fontupper=\scriptsize,
    left=1pt,
    right=1pt,
    valign=center,
    height=.825\textheight
}
\begin{minipage}[c]{.32\textwidth}
\centering
\begin{tcolorbox}[]
To evaluate summaries, consider the following criteria:

1. **Coherence Score**: Assess whether the summary logically flows and connects the key points of the article. Ensure that the summary presents a clear and consistent narrative without contradictions or abrupt shifts.

2. **Consistency Score**: Verify that the summary accurately reflects the content of the article without introducing new information or omitting crucial details. The summary should maintain the same stance and perspective as the original text.

3. **Fluency Score**: Evaluate the readability and smoothness of the summary. The summary should be grammatically correct, well-structured, and easy to understand.

4. **Relevance Score**: Determine whether the summary captures the essential information and main points of the article. Ensure that every sentence in the summary is pertinent to the article's content and does not include irrelevant details.

Each criterion should be scored on a scale of 1 to 5, with 5 indicating the highest quality in that specific aspect.\\
Now please evaluate the following summary to the article based on the above guideline criteria:

Article:
\{article\}

Summary:
\{summary\}

Please just directly output the scores in a json format.
For example:
```json\\
\{\\
    "article": \{article\},\\
    "summary": \{summary\},\\
    "coherence\_score": <a score between 1 and 5>,\\
    "consistency\_score": <a score between 1 and 5>,\\
    "fluency\_score": <a score between 1 and 5>,\\
    "relevance\_score": <a score between 1 and 5>\\
\}\\
```
\end{tcolorbox}
\subcaption{Forward Prompt}
\end{minipage}
\begin{minipage}[c]{.32\textwidth}
\centering
\begin{tcolorbox}[]
You will be given one summary written for a news article.

Your task is to rate the summary on one metric.

Please make sure you read and understand these instructions carefully. Please keep this document open while reviewing, and refer to it as needed.

Evaluation Criteria:

Coherence (1-5) - the collective quality of all sentences. We align this dimension with the DUC quality question of structure and coherence whereby "the summary should be well-structured and well-organised. The summary should not just be a heap of related information, but should build from sentence to a coherent body of information about a topic."

Consistency (1-5) - the factual alignment between the summary and the summarised source. A factually consistent summary contains only statements that are entailed by the source document. Annotators were also asked to penalise summaries that contained hallucinated facts. 

Fluency (1-5): the quality of the summary in terms of grammar, spelling, punctuation, word choice, and sentence structure.

Relevance (1-5) - selection of important content from the source. The summary should include only important information from the source document. Annotators were instructed to penalise summaries which contained redundancies and excess information.

Evaluation Steps:

1. Read the news article carefully and identify the main topic and key points.\\
2. Read the summary and compare it to the news article. Check if the summary covers the main topic and key points of the news article, and if it presents them in a clear and logical order.\\
3. Assign scores on a scale of 1 to 5, where 1 is the lowest and 5 is the highest based on the Evaluation Criteria.

Please evaluate the following summary:
Source Text:

\{article\}

Summary:

\{summary\}

Evaluation Form (scores ONLY):

- Coherence:\\
- Consistency:\\ 
- Fluency:\\
- Relevance:\\
\end{tcolorbox}
\subcaption{Human-Crafted Prompt}
\end{minipage}
\begin{minipage}[c]{.32\textwidth}
\centering
\begin{tcolorbox}[]
You are a highly specialised AI assistant adept at evaluating the quality of various summaries produced by different AI systems. Your task involves assessing a provided article and its associated summaries, with the objective of determining the most accurate and reliable summary. You are required to conduct a comprehensive evaluation of the summaries based on the following criteria:

a) Coherence: Assess the logical flow and coherence of the summary, ensuring that the information presented is logically structured and makes sense as a whole.

b) Consistency: Evaluate how well the summary aligns with the content of the original article, ensuring that all statements are factually consistent with the article and do not introduce any new information.

c) Fluency: Analyze the grammatical accuracy and readability of the summary, ensuring that it is written in a clear and concise manner.

d) Relevance: Determine the extent to which the summary addresses the main points and key details of the article, ensuring that it provides a comprehensive overview.

Please rate each summary on a scale of 1 to 5 (with 5 being the highest) based on the criteria outlined above.

Your response should be in the following format:

\{\\ 
    "article": <article>,\\
    "summary": <summary>,\\
    "coherence\_score": <coherence\_score>,\\
    "consistency\_score": <consistency\_score>,\\
    "fluency\_score": <fluency\_score>,\\
    "relevance\_score": <relevance\_score> \\
\}\\

article: \{article\}

summary: \{summary\}

\end{tcolorbox}
\subcaption{Inverse Prompt}
\end{minipage}
\caption{Prompts for Qwen on Summeval dataset.}
\end{figure}

\begin{figure}[!htb]
\centering
\tcbset{
    colback=gray!5!white,
    fontupper=\scriptsize,
    left=1pt,
    right=1pt,
    valign=center,
    height=.875\textheight
}
\begin{minipage}[c]{.32\textwidth}
\centering
\begin{tcolorbox}[]
To evaluate model responses, consider the following criteria:

1. **Naturalness Score**: Assess whether the response sounds natural and fluent, without awkward phrasing or forced connections. Responses should flow smoothly and be easily understood by humans.

2. **Coherence Score**: Evaluate how well the response aligns with the conversation history and the provided fact. The response should logically follow from the previous exchanges and integrate the given information meaningfully.

3. **Engagingness Score**: Determine whether the response keeps the conversation interesting and engaging. The response should add value to the dialogue, provide relevant information, or provoke further discussion.

4. **Groundedness Score**: Assess whether the response is grounded in the provided conversation history and fact. The response should be relevant and not introduce unrelated or irrelevant information.
Now please evaluate the following model's response according to the conversation and fact based on the above guideline criteria:

Conversation History:
\{conversation\}

Corresponding Fact:
\{fact\}

Response:
\{response\}

Please just directly output the scores in a json format.
For example:
```json\\
\{\\
    "conversation": \{conversation\},\\
    "fact": \{fact\},\\
    "response": \{response\},\\
    "naturalness\_score": <a score between 1 and 3>,\\
    "coherence\_score": <a score between 1 and 3>,\\
    "engagingness\_score": <a score between 1 and 3>,\\
    "groundedness\_score": <a score between 1 and 3>\\
\}\\
```
\end{tcolorbox}
\subcaption{Forward Prompt}
\end{minipage}
\begin{minipage}[c]{.32\textwidth}
\centering
\begin{tcolorbox}[]
You will be given one conversation history and the corresponding facts.

Your task is to rate the conversation history and the corresponding facts on one metric.

Please make sure you read and understand these instructions carefully.

Evaluation Criteria:

Naturalness (1-5) - In order to thoroughly evaluate a model's response according to the conversation history and the corresponding facts, you are required to rate the naturalness of the model's response on a scale of 1 to 5, where 1 indicates very unnatural and 5 indicates very natural.

Coherence (1-5) - the collective quality of all sentences. We align this dimension with the DUC quality question of structure and coherence whereby "the conversation should be well-structured and well-organised. The conversation should not just be a heap of related information, but should build from sentence to a coherent body of information about a topic."

Engagingness (1-5) - engagingness is defined as how effectively the response captures and maintains the interest of the listener or conversational partner

Groundedness (1-5) - groundedness measures the level of factual consistency and relevance in a conversation, ensuring that responses are accurate, contextually appropriate, and supported by reliable sources.

Evaluation Steps:

1. Read the conversation history and the corresponding facts carefully and identify the main topic and key points.\\
2. Read the conversation history and the corresponding facts. Check if the conversation covers the main topic and key points of the corresponding facts, and if it presents them in a clear and logical order.\\
3. Assign scores on a scale of 1 to 5, where 1 is the lowest and 5 is the highest based on the Evaluation Criteria.

Example:

Conversation:

\{conversation\}

Corresponding Facts:

\{fact\}

Evaluation Form (scores ONLY):

- Naturalness:\\
- Coherence:\\
- Engagingness:\\
- Groundedness:\\
\end{tcolorbox}
\subcaption{Human-Crafted Prompt}
\end{minipage}
\begin{minipage}[c]{.32\textwidth}
\centering
\begin{tcolorbox}[]
You are now a conversation evaluation specialist. You will be provided with a conversation history and a fact. You will evaluate a model's reponse to the conversation history on a scale of 1 to 3. The model's response should be natural, coherent, engaging, and grounded in the given fact.

To rate the model's response, you should use the following guidelines:

Naturalness: The response should be natural and sound like something a human would say. It should use appropriate language, tone, and style. The response should not sound like a list of facts or information.\\
Coherence: The response should be coherent and make sense in the context of the conversation history. It should be logically consistent and well-structured.\\
Engagingness: The response should be engaging and keep the conversation interesting. It should include elements that capture the listener's attention and maintain their interest.\\
Groundedness: The response should be grounded in the given fact. It should be relevant to the fact and use it to provide a meaningful and informative answer.\\
You should evaluate the model's response on a scale of 1 to 3 for each of the above criteria. The format of your answer should be as follows:\\
```\\
\{\\
  "conversation": "conversation history", \\
  "fact": "fact", \\
  "response": "model's response", \\
  "naturalness\_score": score, \\
  "coherence\_score": score, \\
  "engagingness\_score": score, \\
  "groundedness\_score": score \\
\}\\
```\\
Where `conversation` is the conversation history, `fact` is the given fact, `response` is the model's response, and `naturalness\_score`, `coherence\_score`, `engagingness\_score`, and `groundedness\_score` are the scores for each of the evaluation criteria. The scores should be a real number between 1 and 3. A score of 1 indicates the lowest level of performance, and a score of 3 indicates the highest level of performance.\\
The conversation history and fact will be provided in the following section.\\
Please read the evaluation guideline carefully and consider it when evaluating the model's response.

conversation history: "\{conversation\}"

fact: "\{fact\}"

model's response: "\{response\}"
\end{tcolorbox}
\subcaption{Inverse Prompt}
\end{minipage}
\caption{Prompts for Qwen on Topical-Chat dataset.}
\end{figure}

\begin{figure}
\centering
\tcbset{
    colback=gray!5!white,
    fontupper=\scriptsize,
    left=1pt,
    right=1pt,
    valign=center,
    height=.775\textheight
}
\begin{minipage}[c]{.32\textwidth}
\centering
\begin{tcolorbox}[]
To evaluate the quality of the machine translation, consider the following criteria:

1. **Accuracy**: The translation should accurately convey the meaning of the original sentence without adding, omitting, or altering information.\\
2. **Fluency**: The translated sentence should read naturally in the target language, maintaining proper grammar, syntax, and word order.\\
3. **Consistency**: The translation should be consistent in terms of tense, person, and number across the sentence.\\
4. **Relevance**: The translation should be relevant to the context and purpose of the original sentence.\\
5. **Preservation of Style**: The translation should maintain the style and tone of the original sentence, including any formal or informal elements.\\
6. **Proper Names and Terms**: Names, titles, and technical terms should be correctly transliterated or translated as per the reference.\\
7. **Punctuation and Spacing**: The use of punctuation and spacing should be correct and consistent with the target language standards.\\
8. **Coherence**: The translation should be coherent and logically connected, ensuring that the sentence makes sense as a whole.

Assign a score from 0 to 100 based on the overall quality of the translation, considering the above criteria.\\
Now please evaluate the following translation based on the above guideline criteria:

Original:\\
\{original\}

Reference:\\
\{reference\}

Translation:\\
\{translation\}

Please just directly output the quality score in a json format.\\
For example:\\
```json\\
\{\\
    "original": "\{original\}",\\
    "reference": "\{reference\}",\\
    "translation": "\{translation\}",\\
    "quality\_score": <a score between 0 and 100>\\
\}\\
```
\end{tcolorbox}
\subcaption{Forward Prompt}
\end{minipage}
\begin{minipage}[c]{.32\textwidth}
\centering
\begin{tcolorbox}[]
Score the following translation from English to German with respect to the human reference on a continuous scale from 0 to 100, where score of zero means "no meaning preserved" in terms of accuracy, contextual understanding, grammar, syntax, overall readability  and score of one hundred means "perfect meaning and grammar" in terms of accuracy, contextual understanding, grammar, syntax, overall readability \\
English source: \{original\} \\
German human reference: \{reference\} \\
German translation: \{translation\} \\
Score:
\end{tcolorbox}
\subcaption{Human-Crafted Prompt}
\end{minipage}
\begin{minipage}[c]{.32\textwidth}
\centering
\begin{tcolorbox}[]
Please assume the role of a highly skilled professional translator. Your task is to evaluate the quality of a machine translation (MT) of a German sentence by comparing it with a reference translation. You are required to rate the quality of the MT on a scale of 0 to 100, where 0 indicates the translation is not useful at all, and 100 indicates the translation is identical to the reference translation. The original sentence and the reference translation are provided below. Note that the machine translation may not be in German, and it may contain grammatical errors or unnatural phrasing. You must evaluate the translation quality in a professional and objective manner, using the format below:\\

\{\\
    "original": "the original sentence", \\
    "reference": "the reference translation", \\
    "translation": "the machine translation", \\
    "quality\_score": a score between 0 and 100\\
\}\\

Ensure that your evaluation strictly adheres to the following guidelines:

1. Do not respond with anything other than the required JSON.\\
2. The quality\_score must be a numerical value between 0 and 100.\\
3. Do not provide any additional information or explanations.\\
4. Compare the machine translation with the reference translation and assess the quality of the translation.\\
5. Consider the meaning, grammar, and fluency of the translation when evaluating the quality.\\
6. Ensure that the evaluation is professional and objective, reflecting the quality of the translation accurately.\\

Original:\\
\{original\}

Reference:\\
\{reference\}

Translation:\\
\{translation\}

\end{tcolorbox}
\subcaption{Inverse Prompt}
\end{minipage}
\caption{Prompts for Qwen on WMT22-EN-DE dataset.}
\end{figure}

\begin{figure}
\centering
\tcbset{
    colback=gray!5!white,
    fontupper=\scriptsize,
    left=1pt,
    right=1pt,
    valign=center,
    height=.5\textheight
}
\begin{minipage}[c]{.32\textwidth}
\centering
\begin{tcolorbox}[]
You are a highly experienced journalist and meticulous fact-checker. Your task is to meticulously evaluate the factual consistency of the summaries provided by other AI systems (such as GPT-3) with the detailed content of the news article. Please note that these summaries may contain factual inaccuracies, such as hallucinations (fabricated details not present in the article) or factual misrepresentations (incorrect information presented as fact). Utilize the following detailed evaluation guideline to score the summaries:

1. A summary is considered to be factually consistent if it does not contain any hallucinated or factually incorrect details. Every detail mentioned in the summary must be explicitly stated in the news article.\\
2. Assign a score between 0 and 1, where a higher score indicates a higher degree of factual consistency.\\
3. The format of the input will be as follows: \{\{article\}\}, \{\{summary\}\}, and the output should be a single score reflecting the factual consistency.\\
article:\\
\{article\}

summary:\\
\{summary\}

score:
\end{tcolorbox}
\subcaption{1 Decimal Place}
\end{minipage}
\begin{minipage}[c]{.32\textwidth}
\centering
\begin{tcolorbox}[]
Please undertake the evaluation of factual consistency of the summaries to the article. I will provide you with a detailed evaluation guideline and a specific format requirement. Your task is to assess whether the summary accurately reflects the information presented in the article, and then assign a numerical score based on the following criteria:

1. Summaries must be factually consistent with the article. A summary is deemed factually consistent if all the numbers in the summary are presented the same way as in the article and if the facts presented align accurately with those in the article.\\
2. You should score the factual consistency of the summary on a scale of \darkred{1 to 10}, where 1 indicates completely inconsistent and 10 indicates completely consistent.\\
3. The input will follow the format provided below, with the article, summary, and a placeholder for the score.\\
4. Evaluate the factual consistency strictly based on the information provided in the article. Do not incorporate any external information.

Here is the format of the input:

\{
    "consistency\_score": score
\}

The article:\\
\{article\}

The summary:\\
\{summary\}

What is the factual consistency score of the summary with the article?
\end{tcolorbox}
\subcaption{w/o Score Range}
\end{minipage}
\begin{minipage}[c]{.32\textwidth}
\centering
\begin{tcolorbox}[]
As an esteemed fact-checker, your responsibility is to meticulously evaluate the factual consistency of the provided summaries in relation to the corresponding articles. Below, you will find the detailed evaluation guideline along with the required format.

Detailed Evaluation Guideline:\\
- Summaries must strictly adhere to the facts presented in the article.\\
- Any inference or speculation not explicitly mentioned in the article must be identified as inconsistent.\\
- The consistency score should be a numerical value ranging from \darkred{0 to 100}, where 100 indicates perfect factual alignment.\\
Output Format:\\
- Your output should be a JSON object with the keys: \"article\", \"summary\", and \"consistency\_score\".

Proceed with your evaluation using the given article and summary. \\
Article:\\
\{article\}

Summary:\\
\{summary\}

Consistency Score:
\end{tcolorbox}
\subcaption{w/o Score}
\end{minipage}
\caption{Prompts for the numerical sensitivity study using Qwen.}
\label{fig:ablation_study_prompts}
\end{figure}
\clearpage
\twocolumn

\section{Meta-Prompt Examples}
\label{app_output}
Here are the meta-prompts (see \S\ref{sec:inversion_prompting}) used for \texttt{Inverse-Qwen} in the Black-Box setting on different tasks to inversely generate the prompt template.
\tcbset{
    colback=gray!5!white,
    fontupper=\scriptsize,
    left=1pt,
    right=1pt,
    valign=center
}

\begin{tcolorbox}[adjusted title={Summarisation},
         colframe=black]

Based on the detailed evaluation guideline and format requirement you provided, I'm now evaluating consistency of the following summaries to the articles with a score between 0 and 1:

```json\\
\{\\
  "article": "A woman who was allegedly raped and abused by eight men in rotherham changed from a `` lovely girl to an animal '' , her mother told jurors . The witness also said her family had been forced to move to spain to escape her daughter 's alleged abusers . Sheffield crown court also heard how police lost tapes of an interview with defendant sageer hussain in 2003 . Eight men , including mr hussain , deny sexually abusing three girls between 1999 and 2003 . The mother of one of the alleged victims said in a statement : `` her character changed from a lovely girl to an animal . She became horrible . '' She said at one stage she discovered a mobile phone in her daughter 's bedroom and rang a number stored under the name'waleed ' . She said a man picked up the phone and said `` i ai n't done owt , i ai n't touched her . It is n't me '' . When she asked her daughter about the phone she said she burst into tears and said `` they 're raping me , they 're raping me '' . She told the court after her daughter went to the police in 2003 her family were repeatedly threatened . `` we were so distraught that we sold the business and the home and moved to spain , '' she said . Det con andy stephanek , of south yorkshire police , told the court the force had lost the tape of an interview with mr hussain when he was first questioned about the allegations . He said it appeared that `` due to the passage of time they 've been destroyed '' . The trial continues .",\\
  "summary": "The mother of a girl accused of being sexually abused by a gang of men has told a court her daughter 's character changed from `` a lovely girl to an animal '' .",\\
  "consistency\_score": 0.6666666666666666\\
\}\\
```
\end{tcolorbox}
\begin{tcolorbox}[adjusted title={Conversation},
         colframe=black]
Thank you for providing a detailed evaluation guideline.
Considering the given fact, I will now evaluate a model's reponse to the conversation history on a scale of 1 to 3.
According to the format requirement, my answer is as follows:

```json\\
\{\\
  "conversation": "hi , do you know much about the internet ? \\ i know a lot about different sites and some website design , how about you ? \\
",\\
  "fact": "the 3 horizontal line menu on apps and websites is called a hamburger button.",\\
  "response": "yeah , i have heard that before . do you know what the hamburger button is?",\\
  "naturalness\_score": 1.6666666667,\\
  "coherence\_score": 2.0,\\
  "engagingness\_score": 2.0,\\
  "groundedness\_score": 1.0\\
\}\\
```
\end{tcolorbox}
\begin{tcolorbox}[adjusted title={Translation},
         colframe=black]

Thank you for providing a detailed evaluation guideline.
Comparing with reference translation, I will now evaluate the machine translation to the original sentence in terms of quality on a scale of 0 to 100.
According to the format requirement, my answer is as follows:

```json\\
\{\\
  "original": "Under the reign of the Yongle Emperor, Ming China underwent a militaristic expansionism with ventures such as the treasure voyages.",\\
  "reference": "Unter der Herrschaft des Kaisers Yongle erlebte China in der Ming-Dynastie einen militaristischen Expansionismus mit Unternehmungen wie Reisen auf der Suche nach Schätzen .",\\
  "translation": "Unter der Herrschaft des Yongle Kaisers erlebte Ming China einen militaristischen Expansionismus mit Unternehmungen wie den Schatzreisen.",\\
  "quality\_score": 95.33\\
\}\\
```
\end{tcolorbox}

\section{Environment details}
\label{app_environment}
We conducted all the training and inference on 4 NVIDIA A100-SXM-80GB GPUs. All the inverse models except for Qwen-2.5-14B are trained for 3 epochs with a total batch size of 1024 (4 devices $\times$ 8 instances per device $\times$ 32 gradient accumulation steps). Due to the VRAM limitation, when training the 14B model, we lower the number of instances per device to 4 and increase the gradient accumulation steps to 64 in order to keep the total batch size the same as 1024. LLaMA-Factory is used \cite{zheng2024llamafactory} as our codebase.
Here is a list of values we set for hyperparameters in the training and inference process:
\begin{table}[!h]
    \centering
    \setlength{\tabcolsep}{5pt}
    \begin{tabular}{lc}
        \toprule
        \textbf{Name} & \textbf{Value} \\
        \midrule
        Total Batch Size & 1024 \\
        Epoch & 3\\
        Learning Rate & 1e-4 \\
        Cutoff Length & 2048 \\
        LoRA Alpha & 512 \\
        LoRA Rank & 256 \\
        \midrule
        Temperature & 0.95\\
        Top P & 0.7\\
        Top K & 50\\
        \bottomrule
    \end{tabular}
    \caption{List of hyperparameters.}
    \label{tab:hyperparameters}
\end{table}

\end{document}